%% file: acl_latex.tex
\documentclass[11pt]{article}

\usepackage[final]{acl}

\usepackage{times}
\usepackage{latexsym}

\usepackage[T1]{fontenc}

\usepackage[utf8]{inputenc}

\usepackage{microtype}

\usepackage{inconsolata}

\usepackage{graphicx}
\usepackage{amsmath}
\usepackage{subcaption}
\usepackage{amssymb}
\usepackage{booktabs}
\usepackage{fontawesome} %
\usepackage[most]{tcolorbox}
\usepackage{enumitem} %

\newcommand{\inlineicon}[1]{\raisebox{-0.2em}{\includegraphics[height=1.1em]{#1}}}
\newcommand{\githubicon}{\inlineicon{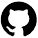}}
\newcommand{\hficon}{\inlineicon{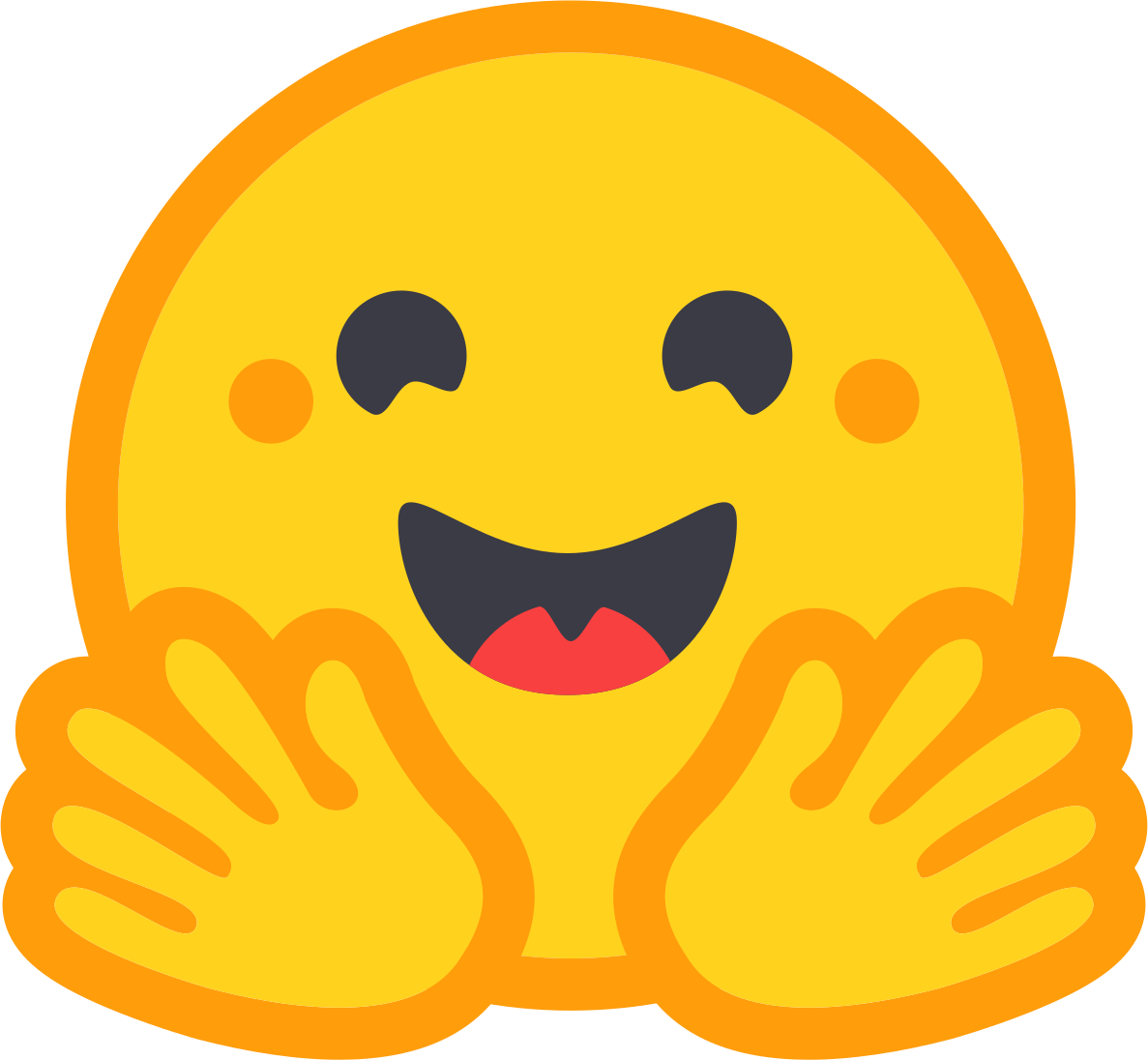}}
\newcommand{\mtebicon}{\inlineicon{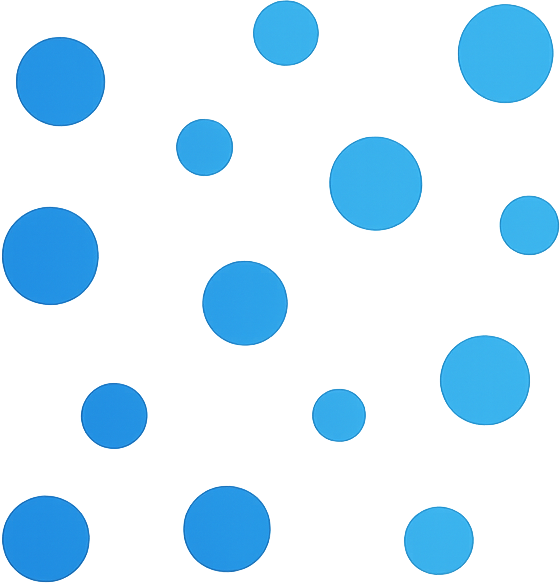}}

\newcommand{\mainresearchquestion}[1]{%
  \begin{tcolorbox}[
    colback=gray!5,
    colframe=black,
    boxrule=0.8mm,
    arc=1.5mm,
    left=2mm,
    right=2mm,
    top=2mm,
    bottom=2mm,
    enhanced
  ]
    {\fontsize{14}{22}\selectfont \faQuestionCircle}\hspace{2mm}%
    \textbf{Research Question:} \textit{#1}
  \end{tcolorbox}%
}

\newcommand{\figref}[1]{Fig.~\ref{#1}} %
\newcommand{\secref}[1]{§~\ref{#1}} %
\usetikzlibrary{arrows.meta,positioning,calc}
\title{Grounding Text Embeddings in Stakeholder Associations}

\author{
  \textbf{Jonathan Rystrøm\textsuperscript{1}},
  \textbf{Sofie Burgos-Thorsen\textsuperscript{2}},
  \textbf{Zihao Fu\textsuperscript{3}},
  \textbf{Johan Irving Søltoft\textsuperscript{4}},
  \\
  \textbf{Kenneth C. Enevoldsen\textsuperscript{5}},
  \textbf{Chris Russell\textsuperscript{1}}
\\
\\
  \textsuperscript{1}University of Oxford, UK,
  \textsuperscript{2}Institute for Wicked Problems, Denmark,
  \\
  \textsuperscript{3}The Chinese University of Hong Kong, China,
  \textsuperscript{4}Danish Technical University, Denmark,
  \\
  \textsuperscript{5}Aarhus University, Denmark
\\
  \small{
    \textbf{Correspondence:} \href{mailto:jonathan.rystrom@oii.ox.ac.uk}{jonathan.rystrom@oii.ox.ac.uk}
  }
  \\[4pt]
  {\small
    \githubicon\ \href{https://github.com/jhrystrom/human-grounding}{\texttt{jhrystrom/human-grounding}} \quad
    \hficon\ \href{https://huggingface.co/datasets/ryzzlestrizzle/stakeholder-grounding}{\texttt{ryzzlestrizzle/stakeholder-grounding}} \quad
    \mtebicon\ \href{https://leaderboard.mteb.org/tasks/StakeholderGrounding}{\texttt{StakeholderGrounding}}
  }
}

\input{latex/key-values}

\begin{document}
\maketitle

\begin{abstract}
Text embeddings are widely used to analyse large corpora of complex texts. However, it is unclear whether the embeddings capture the same semantic relations as the human experts using them. 
Such representational alignment is essential for valid analyses.
We present the Stakeholder Grounding Exercise, a method for making expert associations explicit and grounding embedding model results in human understanding. 
In our primary case study on Danish policy issues, we find that neural text embeddings are substantially less reliable than human experts (\HeadlineGapMinimum-\HeadlineGapMaximum pp gap), and that this misalignment propagates to downstream clustering performance (Spearman $\rho=\GroundingClusteringCorrelation$ between exercise ranking and cluster quality). 
A secondary study on US Federal AI use cases replicates the gap (\GovAIModelGap pp) in English, using a digital protocol and a different community of experts -- demonstrating that the gap is not an artefact of a single instrument or domain. 
The Stakeholder Grounding Exercise offers a practical method for assessing whether embedding models capture the semantic distinctions that matter most to domain experts.
\end{abstract}

\section{Introduction} \label{sec:intro}
Neural text embeddings, such as sentence transformers \citep{reimersSentenceBERTSentenceEmbeddings2019}, are a key tool for analysing complex documents. Fundamentally, they allow researchers to structure unstructured language. 
From clustering legal documents \citep{viannaTopicDiscoveryApproach2024} to analysing public health discussions \citep{xuUnmaskingTwitterDiscourses2022} embeddings help make unruly data legible. By translating language into vectors, embeddings also offer new ways to inform decision-making on complex policy issues: 
Embeddings help researchers cluster and visualise volumes of qualitative data to render legible the messy, contested terrain that many societal problems occupy \citep{winthernielsenPuzzleStateHow2025}.

\begin{figure}[t]
    \centering
    \includegraphics[width=0.95\linewidth]{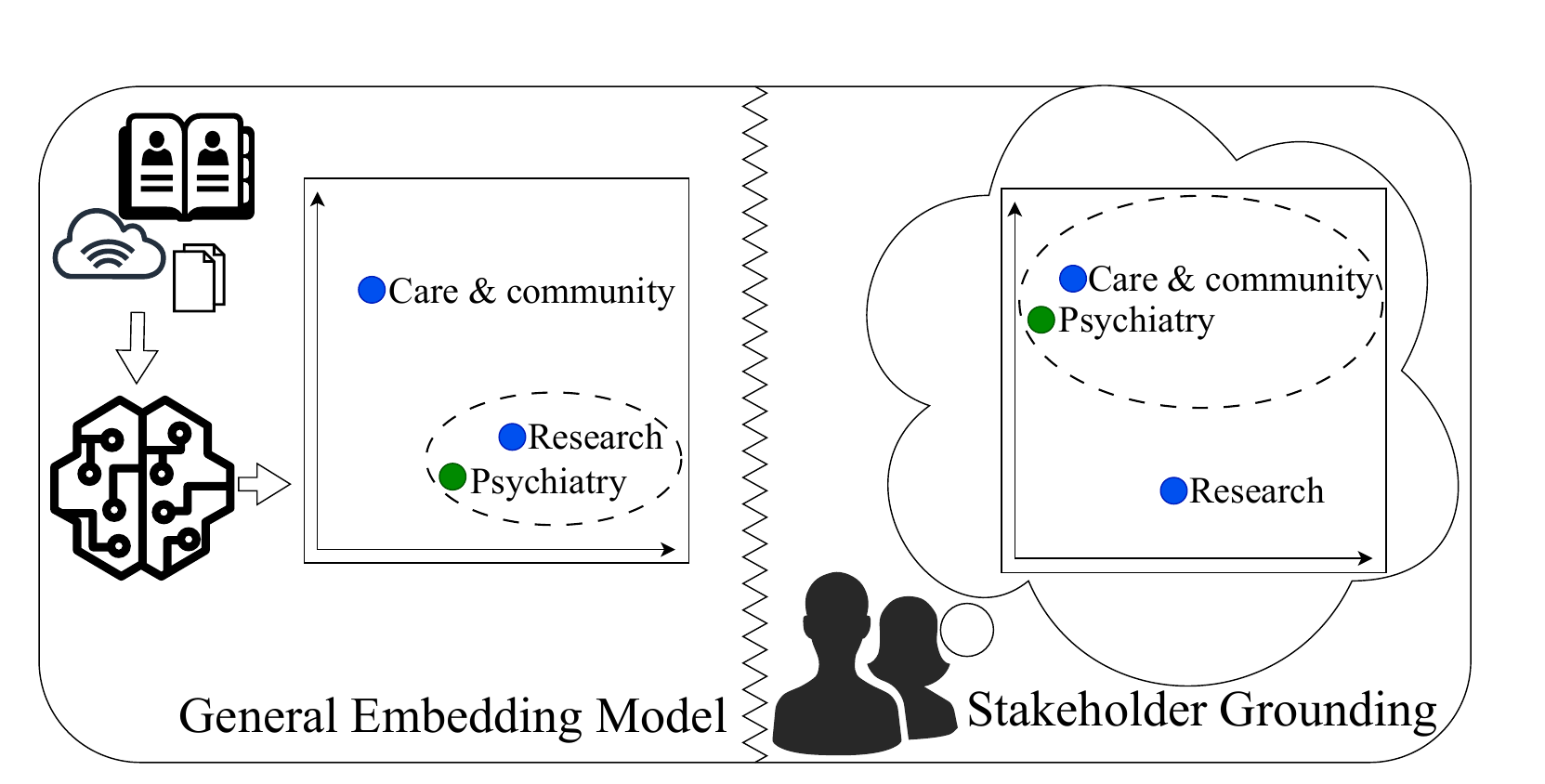}
    \caption{\textbf{The construct gap.} Experts (right) arrange three concepts on a 2D canvas using the Spatial Arrangement Method \citep[SpAM;][]{goldstoneEfficientMethodObtaining1994}, a validated similarity-elicitation paradigm; a neural model embeds them in high-dimensional space (left, schematic). Dashed circles mark closest point (blue) to pivot (green). Current embedding benchmarks cannot detect mismatches; the Stakeholder Grounding Exercise (\secref{sec:human-embedding}) makes them measurable.}
    \label{fig:construct-validity}
\end{figure}

However, the power of these methods depends on a critical assumption: that the representations in embedding space reflect the conceptual distinctions that human experts intend to capture. In practice, this assumption is rarely tested. As \citet{friedlerImpossibilityFairnessDifferent2021} note, models often blur the boundary between the \emph{construct space} -- the theoretical concepts researchers aim to measure -- and the \emph{observed space} derived from data. Text embedding models trained for generic tasks on large-scale data might not match the context-specific constructs of the human analysts. For representation tasks, this gap is especially acute: researchers currently lack systematic ways to compare their intended feature space with that of the embedding model (\figref{fig:construct-validity}).

When researchers cannot examine how well model representations match their intended constructs, they face a \emph{hermeneutical gap} \citep[a conceptual gap in sense-making][]{frickerHermeneuticalInjustice2007}: they lack the means to interpret how models represent meaning \citep{lassenSilencingDataScience2025}. 
This hermeneutical gap has potential downstream impacts; if analyses of the public's input rests on misaligned representations, some perspectives may be inadequately represented \citep{paulladaDataItsDiscontents2021}. 
For instance, clustering together opposing views can mask important disagreements \citep{ghafouriLovePineapplePizza2024}; if these errors are concentrated within specific demographic groups, the effect would be hard-to-measure indirect discrimination \citep{wachterBiasPreservationMachine2021}.

In this paper, we aim to narrow the hermeneutical gap by introducing a method that grounds the evaluation of model representations in human understanding.

\mainresearchquestion{How well do neural text embeddings reproduce expert associations, and what are the consequences when they don't?}

We propose the \emph{Stakeholder Grounding Exercise}, in which key stakeholders create their own two-dimensional embeddings and compare them with neural embedding models (Fig. \ref{fig:construct-validity}, right). The exercise provides a measurable link between the conceptual space of human experts and the feature space of text embeddings. 

We conduct the exercise in two studies of expert sense-making in policy domains. Our primary study examines complex Danish policy issues across two datasets (Responsible AI and the Future of Welfare), where domain experts conduct the exercise on a physical canvas during a workshop. A secondary study on US Federal Agency AI use cases re-runs the exercise in English, using a digital web-app with a community of AI-and-society researchers. 
The two studies share the same exercise and analysis design but differ deliberately in language, instrument, and expert composition, letting us separate findings about the \emph{general exercise} from findings about specific \emph{versions}.

We organise our investigation around three research questions, each yielding a corresponding contribution. 
\textbf{(RQ1)} Does the Stakeholder Grounding Exercise produce valid measurements of stakeholder similarity judgments (\secref{sec:rq1})?
We introduce a reusable \textbf{method} built on the SpAM paradigm \citep{goldstoneEfficientMethodObtaining1994,richieSpatialArrangementMethod2020} and show it is reliable. \textbf{(RQ2)} How reliably do embedding models reproduce those judgments (\secref{sec:rq2})? 
A persistent gap separates the best models from both our expert stakeholders and the \OracleName{} -- one that existing benchmarks do not predict.  
Finally, \textbf{(RQ3)} Do discrepancies in stakeholder grounding affect
downstream clustering (\secref{sec:rq3})? We show that this
misalignment propagates \textbf{downstream} to clustering performance.

\section{Related Work}
\paragraph{Text Embedding Evaluation:}

Text embedding models are commonly evaluated through leaderboards, the most popular of which is MTEB \citep{muennighoffMTEBMassiveText2023,enevoldsenMMTEBMassiveMultilingual2025,chungMaintainingMTEBLong2025}, which spans many tasks, domains, and languages.
This focus on general task rankings leaves limitations. MTEB does not necessarily inform context-specific fairness issues or alignment with groups of human experts \cite{assadiHUMEMeasuringHumanmodel2025}. The Stakeholder Grounding Exercise fills this gap.

\paragraph{Human Evaluation in Embeddings:}

Human evaluation in NLP has traditionally focused on tasks with clear extrinsic objectives such as information retrieval \citep{voorheesTRECExperimentEvaluation2005,resnikEvaluationNLPSystems2010} and classification \citep{liSurveyTextClassification2022}, where humans provide ground truth.

Recently, \citet{assadiHUMEMeasuringHumanmodel2025} compared human and model performance on text-embedding tasks with extrinsic ground truth. They found that humans slightly underperform the best embedding models -- though humans performed better in low-resource languages. Relatedly, \citet{flechasmanriqueEnhancingInterpretabilityUsing2023} use human similarity judgments to prune embeddings in an interpretable way.

Still, these evaluations focus on tasks with clear objectives rather than capturing the associations of domain experts, with whom embedding models can struggle on context-specific distinctions \citep{ghafouriLovePineapplePizza2024}. In contrast, our exercise specifically targets these domain-specific associations, enabling more fine-grained contextual comparisons.

\paragraph{Construct Validity in NLP:}
In social science, `constructs' denote the idealised objects under study, such as `intelligence' or `values' \citep{cronbachConstructValidityPsychological1955}. 
`Construct validity' describes whether the intended constructs are measured appropriately.
Crucially, \citet{friedlerImpossibilityFairnessDifferent2021} showed that a gap between available features and intended constructs has direct fairness implications --- even accurate systems can be discriminatory due to historical data biases \cite{wachterBiasPreservationMachine2021}.

There is an increasing acknowledgement that the evaluation of machine learning systems requires a stronger focus on construct validity \citep{beanMeasuringWhatMatters2025,salaudeenMeasurementMeaningValiditycentered2025}. 
As \citet{bowmanWhatWillIt2021} acknowledge, the focus on beating benchmarks can cause evaluations to drift from meaningful targets. 
\citet{hrytsynaRepresentationResponseAssessing2025} show that model-human judgment can diverge from aggregate benchmark scores.

Our Stakeholder Grounding Exercise aims to catch hard-to-verbalise expert associations in complex domains, following best practices from \citet{jacobsMeasurementFairness2021} and \citet{bowmanWhatWillIt2021}.

\section{Stakeholder Grounding Exercise}  \label{sec:human-embedding}
\subsection{Exercise Description}
\begin{figure}
    \centering
    \includegraphics[width=0.95\linewidth]{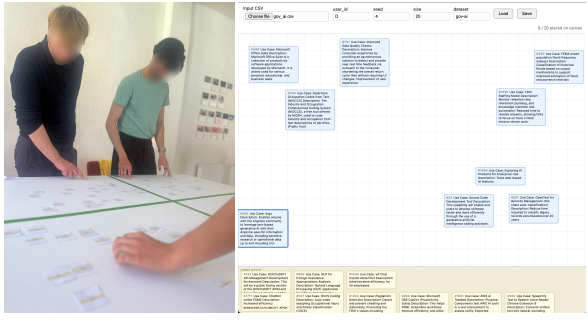}
    \caption{\textbf{Stakeholder Grounding Exercise}. Participants conducting the exercise in-person (left) and online (right). Each round, participants lay out statements on their canvas to represent their notion of similarity.}
    \label{fig:example}
\end{figure}

The Stakeholder Grounding Exercise builds on the \emph{Spatial Arrangement Method} (SpAM; \citealp{goldstoneEfficientMethodObtaining1994}), a free-sorting paradigm developed as an intuitive, time-efficient alternative to pairwise similarity ratings \citep{houtVersatilitySpAMFast2013,houtSpAMConvenientAlso2016} and validated for recovering high-dimensional semantic structure from low-dimensional layouts \citep{kriegeskorteInverseMDSInferring2012,richieSpatialArrangementMethod2020}. In each round, an expert receives 20 statements and arranges them on a 2D canvas so that spatial proximity reflects perceived semantic similarity (Fig.~\ref{fig:example}). The canvas has unlabelled axes: participants are free to choose, round-to-round, the dimensions of variation they find meaningful, allowing the structure of each arrangement to follow expert intuition rather than researcher priors.

Although any single 2D arrangement is a lossy projection, aggregating across rounds and raters recovers the underlying semantic structure, and \citet{richieSpatialArrangementMethod2020} validate SpAM at the $\sim$20-item scale we adopt.
We chose 20 statements per round as a balance between analytic coverage and cognitive load, consistent with evidence that smaller sets reduce fatigue and improve enjoyment \citep{blanchardEvidencebasedRecommendationsDesigning2016}. Statements were sampled uniformly at random, stratified by author demographic group to ensure adequate coverage for fairness analysis (\secref{sec:rq1-methods}). Per-statement occurrence and pair co-occurrence statistics are summarised in Table~\ref{tab:coverage}.
Because rounds sample statements randomly, not every source statement appears, while every selected statement appears repeatedly (mean 3.4-4.0).

In each round, two participants are paired and arrange the same statement set independently, enabling pairwise inter-rater reliability. Because statements recur across rounds, we can also estimate within-rater reliability. We ran two versions of the protocol. For the Danish Policy data, six experts completed 14 rounds each on physical canvases during a half-day workshop (see \secref{sec:data} for dataset description). For Gov-AI, participants used a web-app and completed 7 rounds each in approximately 1--1.5 hours. Both versions yield the round-overlap structure needed for reliability estimation and aggregation across raters.

The output of the exercise is not a single global human embedding; rather we produce local comparisons within each round, and repeated placements across rounds contribute contextual observations. We describe the analysis further in \secref{methods}.

We describe the pre-processing pipeline in Appendix \ref{app:pipeline}. We release all pipelines, the web-app, and the resulting dataset.\footnote{\url{https://github.com/jhrystrom/human-grounding}}

\subsection{Datasets and Protected Attributes} \label{sec:data}
We conduct two studies. The primary study uses two Danish datasets of expert statements on policy problems (`Policy'). A secondary study uses US Federal AI use cases (`Gov-AI'), conducted via a web-app rather than physical canvas. The studies differ along three axes -- language, protocol, and community -- which we use in interpretation.

For Policy, we use two datasets of statements that describe the perceived causes of complex policy problems.  
The first dataset captures a panel of technology practitioners’ perspectives on the challenges of responsibly implementing AI systems (henceforth ``Responsible AI'').  
The second contains Danish local politicians’ statements on the future of welfare (henceforth ``Welfare''). Both datasets contain qualitative text and are mostly in Danish with some English statements. 

The Gov-AI data comes from the US Federal Agency AI Use Case Inventory\footnote{\url{https://github.com/ombegov/2024-Federal-AI-Use-Case-Inventory}}. 
We select the statements on `Mission-enabling' use cases and use the `Intended purpose \& expected benefits' as the key text field. All statements are in English.

To evaluate fairness, we analyse a protected attribute for each dataset's statement authors -- not the exercise participants.  
In Responsible AI, the attribute is \emph{gender}, grouped into two categories: male and female.
In Welfare, the attribute is \emph{political party affiliation} of the politician, pseudonymised with random identifiers to protect respondent anonymity. Political bias in language models is a nascent but fraught area \citep{rottgerPoliticalCompassSpinning2024} with a complex relationship with language \citep{rystromMultilingualMulticulturalEvaluating2025} and important influences on the political process \citep{petersAlgorithmicPoliticalBias2022}.
We include respondents with missing attributes in aggregate analysis but exclude for group-level.
We provide summary statistics for respondents in Appendix \ref{app:respondent-demo}.
Since Gov-AI statements come from agency use cases rather than individual respondents, no protected-attribute fairness analysis applies.

\paragraph{Participants:} We recruited six experts per panel---individuals with substantive familiarity with the domain and prior experience with qualitative similarity judgments in analytical settings. For the Danish policy exercise, five worked at a Danish political think tank; one was an external AI researcher. All held advanced degrees (MSc or PhD), were Danish, aged 25--35, and included four men and two women. 
For the Gov-AI case, all six participants were PhD students working on AI and society from different countries. Full demographics are in Appendix~\ref{app:participants}.

We deliberately recruited a small, domain-specialised panel instead of a demographically broad one. The Stakeholder Grounding Exercise is designed to capture the \emph{interpretative frame} of a community of practice \citep{sloaneParticipationNotDesign2022}. A more heterogeneous panel risks diluting this frame. %

\section{Experimental Methods} \label{methods}
\subsection{RQ1: Human Reliability} \label{sec:rq1}
\subsubsection{Methods} \label{sec:rq1-methods}
Before using the Stakeholder Grounding Exercise to evaluate embedding models, we must establish that it produces valid measurements. 
Following best practices for construct validation \citep{jacobsMeasurementFairness2021, wallachPositionEvaluatingGenerative2025}, we assess three properties: (1)~between-rater reliability, which tests whether the exercise captures a shared expert construct rather than individual idiosyncrasies; (2)~within-rater reliability, which tests measurement stability; and (3)~group-conditional reliability, which tests whether the measurement is consistent across author groups.

We measure all three using a triplet-based agreement measure inspired by 
contrastive learning \citep{khoslaSupervisedContrastiveLearning2020}.  
For each anchor statement, we form all possible pairs of comparison statements and record which one the rater places closer to the anchor in the embedding space.  
Human layouts use Euclidean distance because spatial proximity on the canvas is the core judgment and Euclidean distances are invariant to translation, rotation, and uniform scaling of the arrangement. Model embeddings use cosine distance ($1-\operatorname{sim}_{\cos}$) following standard practice; since embeddings are normalised, this is equivalent to Euclidean distance. We do not compare these values directly; instead, we measure agreement between the induced triplet judgments.
Our agreement measure corresponds to \citet{krippendorffComputingKrippendorffsAlphareliability2011}'s $\alpha$:

\begin{equation} \label{eq:krippendorf}
 \alpha_{\text{Krippendorf}} = 1 - \frac{D_o(\tau)}{D_e}   
\end{equation}

Here, \(D_e\) is the expected random disagreement and \(D_o(\tau)\) is the observed disagreement after filtering triplets by a separation-ratio threshold \(\tau\). The threshold is applied to the observed triplet separation ratio, $r_i(t)$, defined as

\begin{equation}
    r_i(t) = \frac{max\{\delta_i(a,b), \delta_i(a,c)\}}{min\{\delta_i(a,b), \delta_i(a,c)\}},
\end{equation}

where $i$ is the participant's index, $t$ is the triplet of statements $(a,b,c)$ with $a$ being the anchor, and $\delta_i(x,y)$ is the Euclidean distance between statements $x$ and $y$ in participant $i$'s arrangement. Values near 1 represent near ties, while larger values indicate that one statement was unambiguously closer to the anchor. $D_o(\tau)$ is defined in Eq. \ref{eq:disagreement}.

\begin{equation}\label{eq:disagreement}
\begin{aligned}
D_o(\tau) &= \frac{1}{\binom{|\mathcal{R}|}{2}|\mathcal{T}_{i,j}(\tau)|}
\sum_{i < j \in \mathcal{R}}
\sum_{t \in \mathcal{T}_{i,j}(\tau)} \delta_{ij}(t),\\
\delta_{ij}(t) &= \mathbf{1}\{\mathbb{O}_{i}(t) \neq \mathbb{O}_{j}(t)\}.
\end{aligned}
\end{equation}

where \(\mathcal{R}\) is the set of raters, \(\mathcal{T}_{i,j}(\tau)\) is the subset of triplets rated by both \(i\) and \(j\) for which both raters satisfy $r_i(t) > \tau$, and \(\mathbb{O}_i(t)\) is a binary scoring function indicating which item in triplet \(t\) rater \(i\) judges as closer to the anchor.

Since each retained triplet induces a binary judgement---which of two statements is closer to the anchor---we set \(D_e = 0.5\), yielding the equivalent binary form \(\alpha(\tau)=2P(\mathrm{agreement})-1\). 

We primarily compute \emph{between-rater} reliability.  
From these, we compute the total reliability $\alpha^{\text{tot}}$ (average over all triplets), the minimum group-wise reliability $\alpha(\tau)^{\min} = \min_{g} \alpha(\tau)^{g}$, and the maximum group-wise reliability $\alpha(\tau)^{\max} = \max_{g} \alpha(\tau)^{g}$
(pair coverage in Appendix~\ref{app:respondent-demo}). Fairness-wise, $\alpha(\tau)^{\min}$ is a \textit{minimum-rate metric} \citep{mittelstadtUnfairnessFairMachine2024} and $\alpha(\tau)^{\max} - \alpha(\tau)^{\min}$ is a \emph{disparity metric} \citep{mitchellAlgorithmicFairnessChoices2021}.

We expect agreement to rise with higher $\tau$, since these retain only triplets a rater placed unambiguously, filtering out near-ties whose ordering carries little signal.
We evaluate reliability across a range of \(\tau\) values and summarise the results by taking the area under the \(\alpha(\tau)\) curve (AUC). The AUC is computed on a log-spaced grid of thresholds, integrated in log-\(\tau\) space, and normalised by the log span, so that it does not depend on the absolute width of the threshold interval. 
This AUC provides a single reliability score: \(1\) indicates perfect agreement and \(-1\) perfect disagreement. 
Robustness to alternative threshold ranges, grid sizes, and linear-\(\tau\) integration is reported in Appendix~\ref{app:metric-robustness}.

\subsubsection{Results}
\begin{figure}
    \centering
    \includegraphics[width=\linewidth]{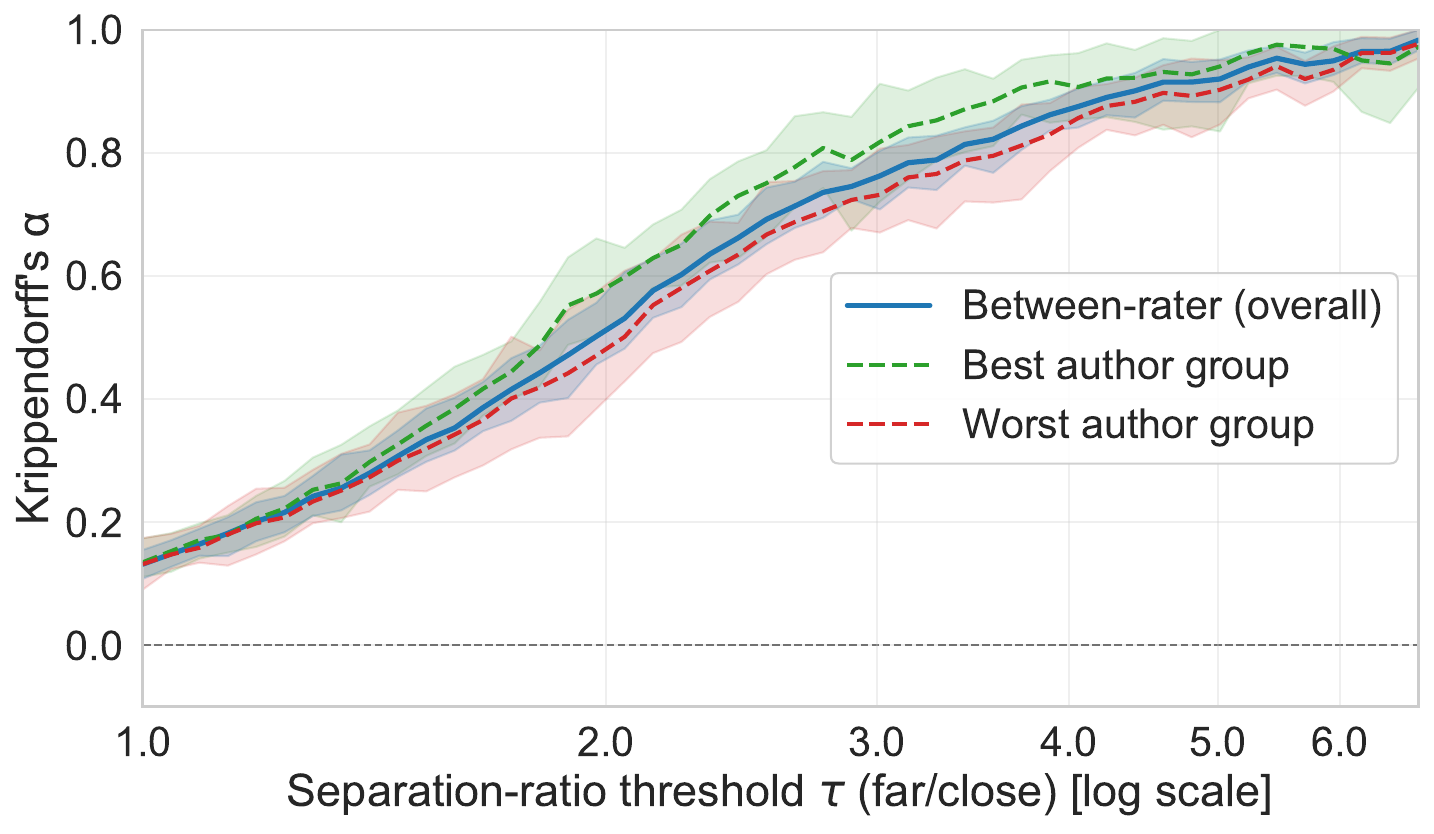}
    \caption{\textbf{Human reliability.} Relationship between the separation-ratio threshold, $\tau$, and inter-rater reliability ($\alpha$). The reliability increases approximately logarithmically with 80\% reliability at $\tau\approx3.1$.}
    \label{fig:rq1-human-reliability}
\end{figure}

Fig.~\ref{fig:rq1-human-reliability} shows the relationship between the separation-ratio threshold, $\tau$, and the interrater reliability measured with \citet{krippendorffComputingKrippendorffsAlphareliability2011}'s $\alpha$. We find an almost monotonic increase in reliability that flattens out as $\alpha$ approaches 1.0. These results indicate that the human analysts agree on much of the data. Across all thresholds, we find a high within-rater agreement and relatively low-levels of contextual drift (see Appendix \ref{app:context-drift}). 
Appendix~\ref{app:respondent-demo} reports retained triplet counts and co-occurrences; even at \(\tau=4\), above the point where human reliability reaches approximately \(\alpha\approx0.8\), every author group retains at least 469 triplets, and the Responsible AI author groups retain more than 1,500.

Finally, the difference in reliability between author groups is relatively minor. The confidence intervals in Fig. \ref{fig:rq2-scores} (human bars) often overlap, and the differences are generally small. For instance, the Responsible AI author groups ranged from $\alpha_{\mathrm{AUC}}^{\text{Men}} = \ResponsibleAIMenAUC\,[\ResponsibleAIMenAUCLO,\,\ResponsibleAIMenAUCHI]$ to $\alpha_{\mathrm{AUC}}^{\text{Women}} = \ResponsibleAIWomenAUC\,[\ResponsibleAIWomenAUCLO,\,\ResponsibleAIWomenAUCHI]$.  Thus, we find no consistent group-wise labelling bias for human annotators.

\subsection{RQ2: Neural-Human Alignment} \label{sec:rq2}
\subsubsection{Methods} \label{sec:rq2-methods}
This research question examines how reliably neural embeddings reproduce human expert embeddings. It has two parts.

First, we identify suitable candidate models using the MMTEB benchmark \citep{enevoldsenMMTEBMassiveMultilingual2025}, which provides detailed metadata on embedding types and languages. We test whether the highest-ranking models on the Danish subset also rank highly in reproducing human embedding structures.
We evaluate 32 models including sentence-transformers \citep{reimersSentenceBERTSentenceEmbeddings2019}, instruction-tuned embedding models \citep[e.g.,][]{wangMultilingualE5Text2024}, commercial embedding models, and Model2Vec \citep{tulkensModel2VecFastStateoftheart2024}. A full list is available in Appendix \ref{app:embedding-list}. 
For the instruction-tuned embedding models, we provide generic instructions to `cluster similar sentences' following their specified formats (see Appendix \ref{app:instruction} for full prompts). 
Across five instruction variants in generic and dataset-context formulations (10 conditions per model; Appendix \ref{app:instruction}, Fig. \ref{fig:instruct-results}), performance varied modestly (max. gain \MaximumInstructionGain{} over default), consistent with \citet{suOneEmbedderAny2023} -- with context helping some model–dataset pairs and hurting others. Exhaustive prompt optimisation and model adaptation are out of scope.
To test whether surface form alone explains groundedness, we include two lexical baselines: character \(3\)--\(5\)-gram TF--IDF and binary word-overlap representations. We list all baselines and embedding models in Appendix~\ref{app:embedding-list}.
As an upper reference, we also compute a full-panel \OracleName{} by applying MDS to the consensus distance matrix (Appendix \ref{app:oracle}).

Second, we construct the evaluation data. We use the same triplet sets and separation-ratio thresholds as in RQ1. For each model, we compute 
$\alpha_{\text{Krippendorff}}$ (Eq.~\ref{eq:krippendorf}) between the model and each human rater, treating the model as an additional rater. 
Model--human scores pool triplet-level model--rater comparisons within each model--dataset--author-group cell before computing \(\alpha(\tau)\).
For each demographic group, the corresponding cell contains all triplets with at least one statement authored by a member of that group, so triplets may contribute to multiple author groups. 
This tests whether statements from different author groups are represented with comparable fidelity.
CIs use a seeded hierarchical bootstrap that resamples raters, and then triplets within raters, holding the rater resample fixed across \(\tau\)-thresholds for each AUC replicate. 
$\tau$ is based on the human rater's $r_i(t)>\tau$.

We summarise each model's performance by taking the area under the $\alpha(\tau)$ curve, yielding a single groundedness AUC. This avoids dependence on any single threshold choice and mirrors the reliability AUC used for humans.

To compare models, we aggregate groundedness scores across datasets using a Borda count, which treats each dataset as a voter for the best model \citep{enevoldsenMMTEBMassiveMultilingual2025}. Average rank handles tie breaks. We also compute the Spearman rank correlation \citep{spearmanProofMeasurementAssociation1904} between MMTEB rankings and human-grounded rankings, since the analysis focuses on rank-wise consistency.

\begin{figure}
    \centering
    \includegraphics[width=0.95\linewidth]{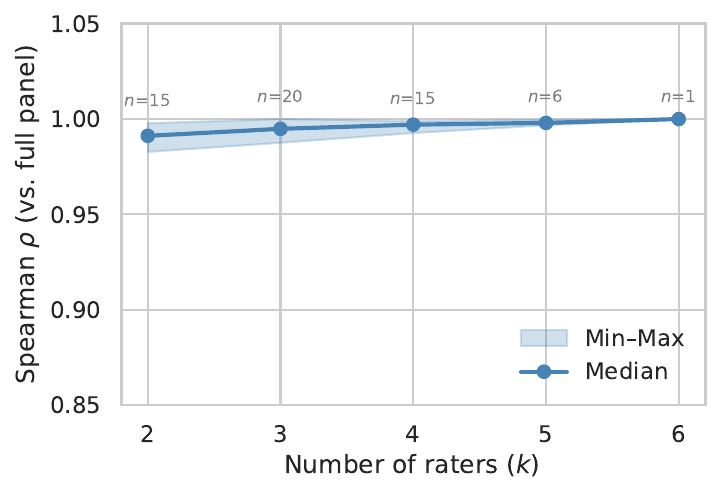}
    \caption{\textbf{Model rank robustness.} Spearman's $\rho$ between rankings from the full six-expert panel and all $\binom{6}{k}$ rater subsets. Points show subset means; shading shows min--max range. Rankings are highly stable across all subset sizes ($\rho > 0.98$). See \ref{sec:rq2-methods} for description.}
    \label{fig:rater-stability}
\end{figure}

Finally, we assess the stability of model rankings with respect to panel size and composition. For each \( k \in \{2,\dots,5\} \), we enumerate all \( \binom{6}{k} \) expert subsets, recompute the Human-Grounded ranking, and measure Spearman’s \( \rho \) against the full six-expert ranking (reference). Rankings remain highly stable across all subset sizes (\( \rho > 0.98 \); Fig.~\ref{fig:rater-stability}), indicating robustness to panel composition despite the small number of experts.

\subsubsection{Results} \label{sec:rq2-results}
\begin{figure}
    \centering
    \includegraphics[width=0.95\linewidth]{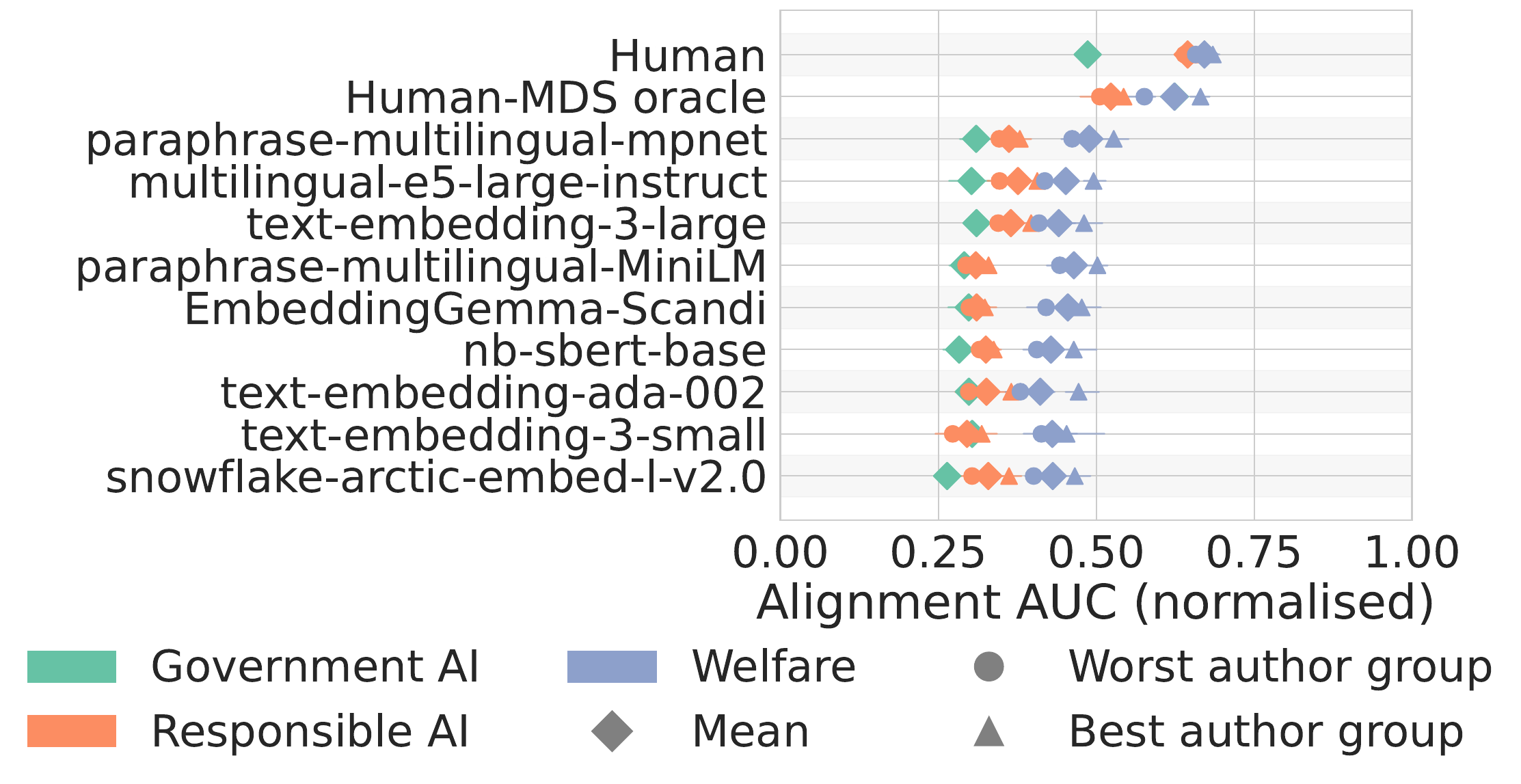}
    \caption{\textbf{Performance gap and disparities}. The best embedding models substantially underperform stakeholder grounding across all datasets. `Human' is between-rater, others are model-human (\secref{sec:rq2}). There are also disparities between author groups.}
    \label{fig:rq2-scores}
\end{figure}

\paragraph{Alignment gap:}
Here, we compare the performance of the 32 different embedding models. The main results are shown in Fig. \ref{fig:rq2-scores}. Even the best embedding models have a significant gap to human inter-rater reliability. The gap is \WelfareModelGap pp (\WelfareModelGapLO-\WelfareModelGapHI pp) for the welfare dataset and \ResponsibleAIModelGap pp (\ResponsibleAIModelGapLO-\ResponsibleAIModelGapHI pp) for the responsible AI dataset.
For the English-language Gov-AI dataset, we see similar differences with a gap of \GovAIModelGap pp (\GovAIModelGapLO-\GovAIModelGapHI pp), indicating that the phenomenon is not language- or dataset-specific.
These differences persist across different difficulties of triplets comparing the top 20\% most vs least challenging triplets (evaluated across $\tau$; see Fig. \ref{fig:difficulty}).
Relative to the \OracleName{}, we also find substantial gaps of \WelfareOracleGap pp (for Welfare), \ResponsibleAIOracleGap pp (for Responsible AI), and \GovAIOracleGap pp (for Government AI). These gaps show that global embeddings \emph{could} represent the stakeholders better.
The \OracleName{} need not match human-human agreement as it produces a single embedding per statement rather than local, contextual arrangements; that it exceeds it for Gov-AI but not for the Policy datasets suggests differing amounts of recoverable structure. 
The recoverable structure may be a result of constraining to two dimensions (see Limitations).
We also find large variation in the inter-model reliability with scores from $-0.4-0.9$ (see Fig. \ref{fig:intermodel}).
These findings also hold when using Spearman correlation instead of our AUC metric (Fig. \ref{fig:rq2-spearman-ablation}).

The best model across all datasets is the \BestEmbeddingModel{} model from sentence-transformers \citep[$\alpha_{\text{AUC}}=\BestEmbeddingModelAUC$;][]{reimersSentenceBERTSentenceEmbeddings2019} followed by the \SecondEmbeddingModel{} \citep{wangMultilingualE5Text2024}. Both are inherently multilingual but have not specifically been fine-tuned for Danish. 

\begin{figure}[htpb]
    \centering
    \includegraphics[width=0.95\linewidth]{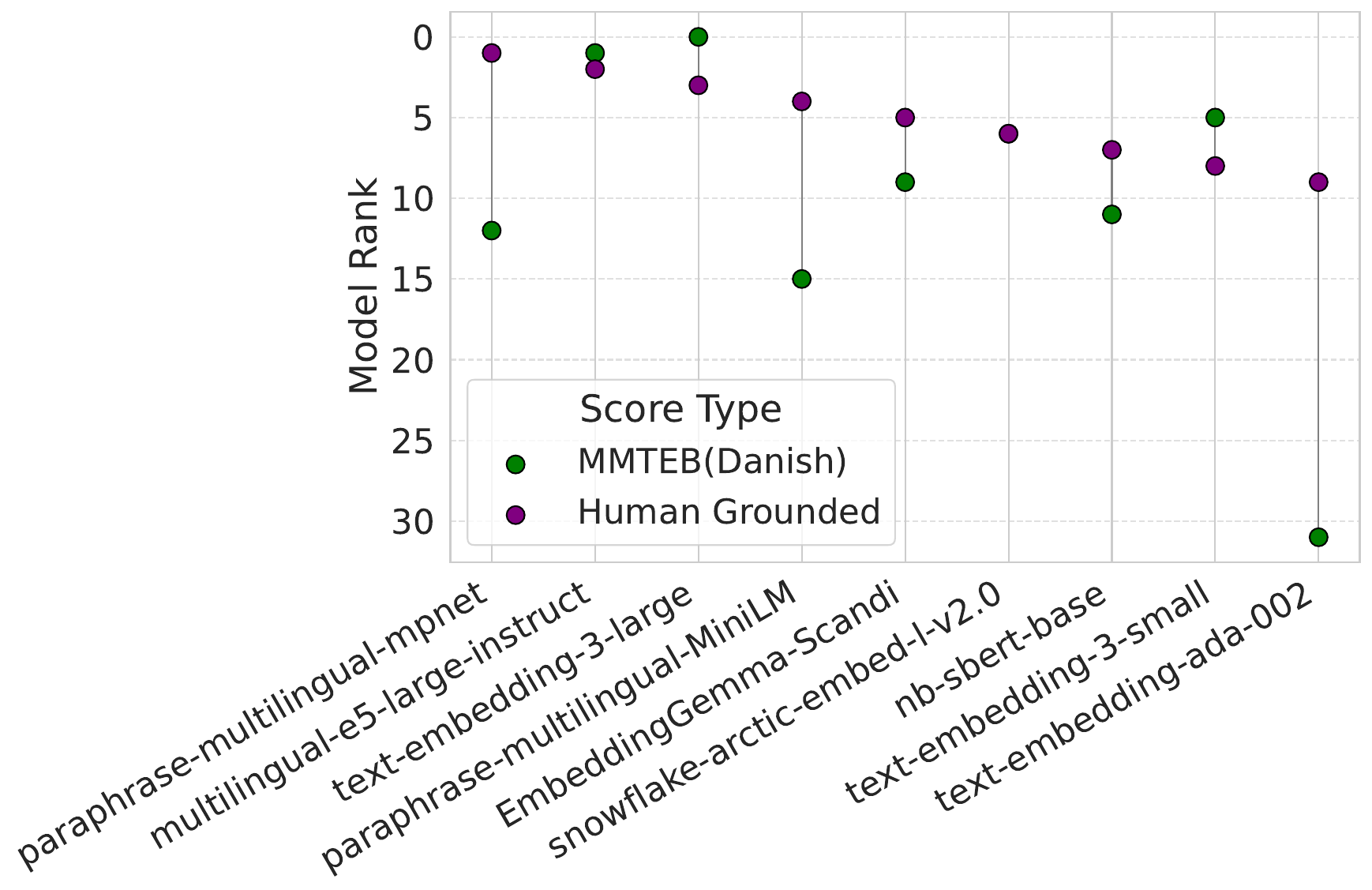}
    \caption{\textbf{Rank comparison for stakeholder grounding and MMTEB.} The best performing models in our stakeholder grounding exercise and Danish subset of MMTEB \citep{enevoldsenMMTEBMassiveMultilingual2025} differ. The Spearman correlation is \MMTEBGroundedSpearman{} ($p=\MMTEBGroundedPValue$).}
    \label{fig:rq2-mmteb-ranking}
\end{figure}

\paragraph{Rank correlation:}
We also compare the ranking of models against the rankings from the Danish subset of MMTEB \citep{enevoldsenMMTEBMassiveMultilingual2025}. These results are in Fig. \ref{fig:rq2-mmteb-ranking}. We find substantial disagreements in ranking. The best model for our task is only ranked 14th on MMTEB. The overall \citet{spearmanProofMeasurementAssociation1904} correlation is \MMTEBGroundedSpearman{} ($p=\MMTEBGroundedPValue$).
This indicates that our exercise captures something different from general ability and that MMTEB is a useful but not complete source for model selection. Specifically, on our datasets, OpenAI's newer text embedding models seem to underperform their MMTEB ranks, whereas some simpler, non-retrieval-trained sentence-transformer models trained on broad, multilingual similarity tasks overperform their ranks. 
However, the performance disparities are not consistent across model groups (see Table \ref{tab:model-type-rank-delta}). The Stakeholder Grounding Exercise mainly helps in selecting individual models.

Furthermore, embedding models exhibit systematic (albeit small) author-conditional disparities (Fig.~\ref{fig:rq2-scores}). For the best models, the disparity is \(\Delta_{\mathrm{group}}=\ResponsibleAIGroupGap\) for Responsible AI (\(95\%\ \mathrm{CI}=[\ResponsibleAIGroupGapLO,\ResponsibleAIGroupGapHI]\)) and \(\WelfareGroupGap\) for Welfare (\([\WelfareGroupGapLO,\WelfareGroupGapHI]\), Appendix~\ref{app:respondent-demo}).
The disparity persists when controlling for statement length and lexical overlap.

\paragraph{Qualitative patterns:}
We qualitatively analyse model-human triplet comparisons with $r_i(t) > 13$ for the best-performing model (\textit{paraphrase-multilingual-mpnet}) to identify where it diverges from expert judgements (full data and original Danish in Appendix \ref{app:triplets} and the repository). Across both policy datasets, the model classifies most high-confidence triplets correctly. When it fails, the errors are systematic: the model relies on surface lexical and topical similarity, whereas experts apply a shared domain-specific interpretative frame, reading statements as \emph{problem diagnoses} within a domain.

The Welfare dataset illustrates this pattern. Experts judge \textit{``Complicated and comprehensive central regulation [...] municipalities lack freer frameworks for local solutions''} as closer to \textit{``It takes too long to revise the Budget Law''} than to \textit{``social inequality and polarisation''}, whereas the model produces the opposite ordering. For experts, the Budget Law is a form of central regulation; both statements diagnose the same institutional constraint. The model, lacking this legal--political frame, defaults to broader topical associations.

The Responsible AI dataset shows a similar pattern. The model groups statements that share AI-harm vocabulary but differ in diagnostic framing. Experts distinguish between \emph{intentional manipulation} (e.g., generating false content), \emph{systemic bias} (e.g., reinforcing historical stereotypes), and \emph{epistemic fragility} (e.g., difficulty fact-checking AI-generated content). These diagnoses imply different causes and policy responses, yet the model collapses them into a single semantic neighbourhood -- sometimes with $r_i(t)>15$.

The same pattern appears with terse statements: \textit{``Education''} is correctly placed via keyword overlap, while the equally terse \textit{``Psychiatry''} is misclassified as experts read it as an institutional diagnosis rather than a topic label.
Consistent with this qualitative interpretation, the lexical baselines underperform most embedding models on both groundedness and downstream clustering (Appendix~\ref{app:embedding-list}), indicating that surface overlap alone does not explain the human structure.

The group-wise gaps in Fig.~\ref{fig:rq2-scores} surface potential representational risk. While the gaps we find are small, the Stakeholder Grounding Exercise makes these asymmetries measurable -- addressing the hermeneutical gap flagged in §\ref{sec:intro}.

\subsection{RQ3: Downstream clustering} \label{sec:rq3}
\subsubsection{Methods} \label{sec:rq3-methods}
Finally, we analyse the downstream differences between neural embeddings and human embeddings. Specifically, we investigate clustering -- a common problem in document analysis \cite{aggarwalSurveyTextClustering2012}. We compare clusters calculated from the human data with text embedding-based clusters. 
While this provides a \emph{downstream} validation, it is not \emph{external} as we lack independent labels. 

To create the human clusters, we use local, per-round agglomerative clustering \citep{wardHierarchicalGroupingOptimize1963}. While text embeddings are generally global, there is increased interest in making them contextual \citep{morrisContextualDocumentEmbeddings2024}. For each round and participant, we cluster together statements within a specified separation-ratio threshold. We construct a test-set of manually labelled clusters to select the threshold. Since two participants see each unique round of statements, we can calculate a human consistency score. We assign a `noise'-label to singular clusters and remove these from the final analysis.

For the neural text embeddings, we use a similar clustering technique \cite{wardHierarchicalGroupingOptimize1963}. We provide the number of clusters as the average number of clusters found by the participants. To compare these clusters with the humans, we take the average adjusted \citet{randObjectiveCriteriaEvaluation1971} index across both participants. 
For each round, a model's clusters are compared with each participant's clustering and the two ARIs averaged; the human score compares the paired participants directly. Because grounding AUC and clustering ARI derive from the same layouts, we interpret clustering as a downstream consequence within this evaluation setting rather than fully independent corroboration.
We analyse the same 32 models (and two baselines) as in \secref{sec:rq2-methods}.

\subsubsection{Results} \label{sec:rq3-results}
\begin{figure}[htbp]
    \centering
    \includegraphics[width=\linewidth]{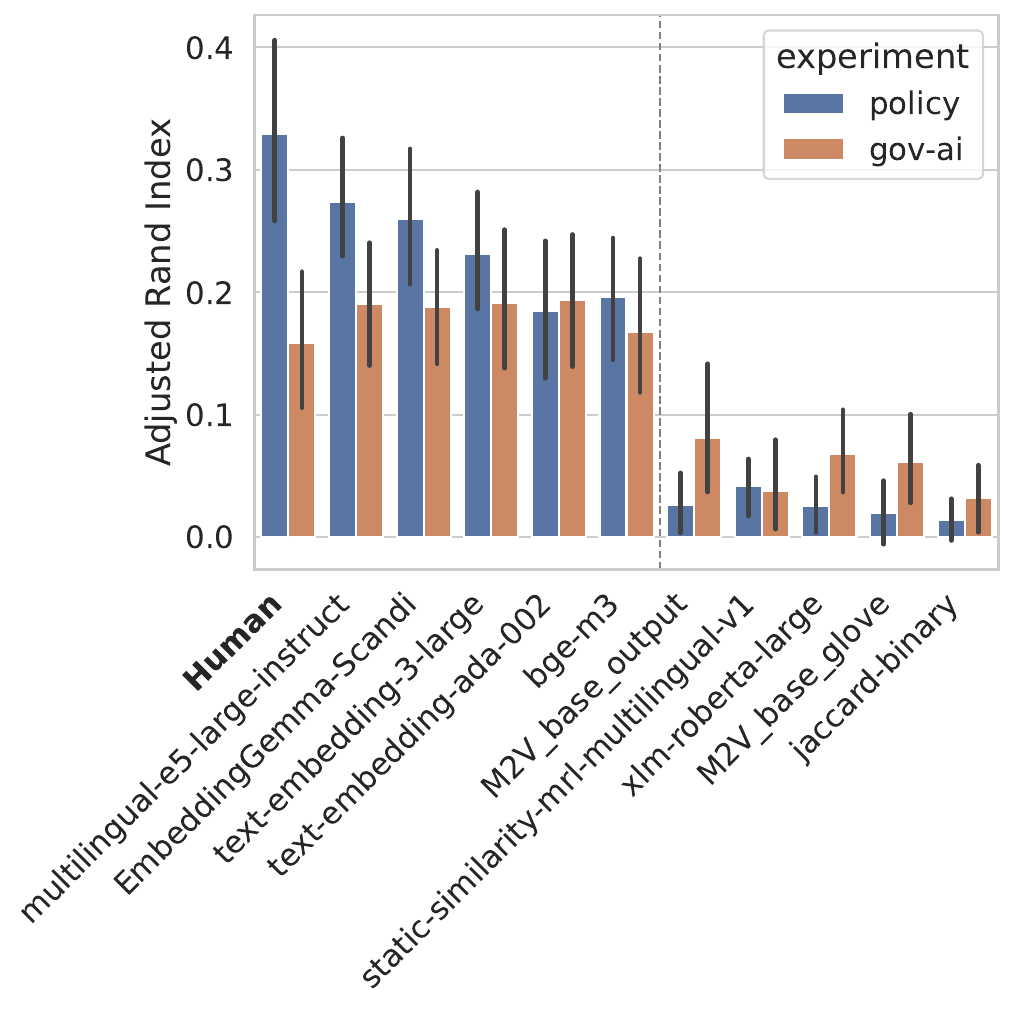}
    \caption{\textbf{Clustering performance}. Comparison of clustering performance (Adjusted Rand Index) for human experts versus the top embedding models. Most embedding models underperform humans. Full results in Table \ref{tab:cluster-consistency-adjusted-rand-index}.}
    \label{fig:rq3}
\end{figure}

Fig. \ref{fig:rq3} shows the Adjusted Rand Index for humans versus models across the datasets. Humans outperform all models in the Danish policy experiment ($\text{ARI}=\PolicyHumanARI\,[\PolicyHumanARILO,\,\PolicyHumanARIHI]$), though not always significantly. For the Gov-AI experiment, humans are mediocre, though not significantly outperformed by any embedding model. Overall, humans rank first on a Borda-rank \citep{mcleanBordaCondorcetPrinciples1990}. We find a high Spearman correlation between the clustering and our stakeholder grounding exercise ($\rho=\GroundingClusteringCorrelation,CI=[\GroundingClusteringCorrelationLO,\GroundingClusteringCorrelationHI]$), which is substantially higher than the correlation with MMTEB ($\rho=\MMTEBClusteringCorrelation,CI=[\MMTEBClusteringCorrelationLO,\MMTEBClusteringCorrelationHI]$) on the Policy datasets.
Results were robust to clustering choices: replacing Ward linkage with k-means yielded highly similar model rankings (Spearman \(\rho = 0.962\), \(p < 0.001\)) and nearly identical human--model ARI gaps (\(\Delta = 0.009\)); results were also stable when selecting \(K\) internally by silhouette score rather than using the human-derived cluster count (Appendix~\ref{app:kmeans}).

The results highlight the tension between \emph{local} and \emph{global} embeddings. The human embeddings are inherently local; each round, the participants arrange a set of statements, influenced by the contextual cues within that specific set. In contrast, the neural embeddings produce \emph{global} representations independent of the surrounding statements. 
This tension is mirrored in real-world use. Social scientists apply these global embeddings to perform contextual analysis of complex constructs \citep[e.g.,][]{xuUnmaskingTwitterDiscourses2022}. 
The qualitative failures identified in \secref{sec:rq2-results}, where models miss contextual cues are the kind of local, contextual associations that global representations cannot capture. 
The Stakeholder Grounding Exercise makes this tension visible and measurable, providing a basis for evaluating whether future contextual approaches \citep[e.g.,][]{morrisContextualDocumentEmbeddings2024} can close the gap.

\section{Conclusion}
We introduced the Stakeholder Grounding Exercise, a method that makes expert associations measurable, providing a bridge between the construct space of human analysts and the feature space of neural text embeddings. Our primary study on Danish policy issues yields three key results, which our Gov-AI study broadly replicates despite differences in language, protocol, and experts.

\textbf{First}, human experts produce reliable embeddings: inter-rater agreement reaches $\alpha \geq 0.8$ when evaluated on triplets with sufficient separation ratio, with no consistent fairness disparities across author groups; the Stakeholder Grounding Exercise makes this visible. 
\textbf{Second}, even the best neural embedding models fall \HeadlineGapMinimum--\HeadlineGapMaximum{} percentage points short of human inter-rater reliability (main experiment), and their rankings diverge substantially from those on MMTEB ($\rho = \MMTEBGroundedSpearman$), indicating that general-purpose benchmarks are insufficient for domain-specific model selection. 
Neither industry nor research practitioners can rely on benchmarks alone; model selection should be supplemented by local, domain-specific human assessment
\cite{loukissasAllDataAre2022}, which the Stakeholder Grounding Exercise can provide.
\textbf{Third}, this misalignment propagates downstream: stakeholder grounding scores predict clustering performance far better than MMTEB rankings ($\rho = \GroundingClusteringCorrelation$ vs.\ $\rho = \MMTEBClusteringCorrelation$).

These findings have practical implications: researchers and domain experts who select embedding models based on leaderboard performance may end up with models that poorly reflect their intended constructs in domain-specific, contextual settings. The Stakeholder Grounding Exercise offers a lightweight, reusable alternative -- grounding model selection in the judgments of the people who will act on the results. We release our code and materials to support adoption in new domains.\footnote{\url{https://github.com/jhrystrom/human-grounding}}

\section*{Limitations}
Our study has two relatively small, homogeneous participant pools ($N_{total}=12$). 
This precludes us from generalising our findings to a broader population. 
However, the goal of the exercise is precisely to capture the associations of a group of experts. We illustrate this across two different contexts -- Danish policy evaluation and AI use in Government. 
The small group of participants allows us to showcase our exercise methodology effectively and acts as a proof-of-concept for other researchers working in resource-constrained settings. 
We also show in Fig. \ref{fig:rater-stability} that our analysis is stable.

While these contexts are different -- one concerns policy the other technology development in government -- it is still an open question whether the gaps generalise to substantially different domains like literature or biology. We welcome researchers to use our methods to answer these questions.

The Danish part of data collection was done in a single half-day workshop. This involves some practical efforts, such as producing the paper statements and gathering people for the workshop. We circumvent this effort using the digital canvas, which we showcased in the Gov-AI use case.

Having only a 2D canvas available constrains the similarity judgments. 
While previous scholarship on related methods shows that it can still recover high-dimensional associations \citep{richieSpatialArrangementMethod2020}, it might still affect our specific exercise. However, this acts as a \emph{conservative} limit to the gap, since the low-dimensionality potentially induces disagreements, which strengthens our finding of a persistent gap, which the \OracleName{}, aggregating across rounds into a higher-dimensional global structure, corroborates empirically.
Ultimately, our exercise must balance \emph{expressiveness} with \emph{complexity}. Informally, participants mentioned that it could be difficult to fit all relations in 2D, but also noted the complexity of getting an overview of all statements -- something that higher dimensions would make even more challenging.
Our Stakeholder Grounding Exercise is complementary to more traditional evaluation methods, not meant as a replacement.

Finally, human arrangements are inherently local and contextual: each layout reflects the set of statements from that round, whereas model embeddings are global and context-free. This is by design -- the exercise elicits contextual stakeholder judgements -- and our within-rater and context-drift analyses (Appendix \ref{app:context-drift}) show that these local judgments are stable. Moreover, the \OracleName{} demonstrates that a single representation per statement can recover much of this structure, so the gap is not attributable to locality alone.

\section*{Ethical considerations}
\paragraph{Human subjects and labour:}
This study was approved by our Institutional Review Board (IRB) prior to data collection. Participants in the \textit{Stakeholder Grounding Exercise} were recruited from the authors' professional network and the partner organisation. Participation was voluntary and conducted as part of the participants' standard professional or academic activities; consequently, no additional financial compensation was provided. All participants provided informed written consent regarding the use and retention of their contributions. To ensure anonymity, we do not disclose specific institutional affiliations in this manuscript.

\paragraph{Data privacy:}
We utilise secondary datasets comprising statements from local politicians and expert survey responses. The political dataset was pseudonymised by the data providers prior to analysis. Given the generic nature of the text segments and the large population of local politicians in the region, we assess the risk of individual re-identification as minor. The expert survey data was collected with consent for secondary research use. 
The Gov-AI dataset comprises public statements by government organisations and thus has no privacy concerns.
We strictly adhere to the data governance protocols established by the data owners.

\paragraph{Demographic limitations:}
We explicitly acknowledge that our panel of human experts is demographically homogenous, aligning with the ``WEIRD'' (Western, Educated, Industrialized, Rich, Democratic) profile \cite{henrichMostPeopleAre2010}. Therefore, the ``stakeholder ground truth'' established in this study reflects a specific socio-cultural and educational standpoint rather than a universal baseline. We caution against generalising these findings to all human judgment. The \textit{Stakeholder Grounding Exercise} should be viewed as a methodology for aligning models with \textit{specific} stakeholder consensus, rather than a claim of solving universal human alignment \citep{sloaneParticipationNotDesign2022}.
Selecting an appropriate panel for a given application is a critical design decision that determines the interpretative frame. This decision must be carefully documented.

\paragraph{Broader impact:}
Our findings indicate that embedding models may misalign with expert consensus, potentially leading to \textit{transactional injustice} \citep[downstream consequences for the disadvantaged group;][]{rauhCharacteristicsHarmfulText2022} if used uncritically in downstream applications. 
Even if such bias stems from technical limitations rather than intent, the consequences for underrepresented groups in the embedding space remain significant. 
Our measurements of group-conditional alignment disparities are statistically robust (Table \ref{tab:group-gap-controlled}), though modest ($\Delta_\text{group} \leq \WelfareGroupGap$). We interpret them as potential indicators of representational risk; whether they translate into consequential transactional injustice depends on the downstream application and the affected group.
By quantifying these misalignments, we aim to prevent the silent replacement of expert nuance with machine approximation.

\section*{Acknowledgments}
We thank the Institute for Wicked Problems (INVI) for making available data, providing participants and workshop space.
A special thanks to Sam Rahbar for helping with practical preparation and ideation.
Also thanks to DPhil students in Social Data Science at the Oxford Internet Institute for participation. JR is supported by the Engineering and Physical Sciences Research Council [Grant Number EP/W524311/1]. KE is funded by Danish Foundation Models (4378-00001B),  the European Union, Horizon Europe (101178170), the Danish National Research Foundation (DNRF193), the Aage and Johanne Louis-Hansens Foundation (25-1-17733), and the Augustinus Foundation (2025-0299).
Claude Code and GitHub Copilot were used for coding assistance. All suggested code and outputs were reviewed by the authors in line with the Guidelines for Generative Assistance in Authorship.

\bibliography{latex/zotero}

\appendix
\section{Coordinate pipeline illustration} \label{app:pipeline}
\usetikzlibrary{arrows.meta,positioning,calc}
\captionsetup[subfigure]{font=small,labelformat=simple,justification=centering}
\begin{figure}[t]
  \centering
  \begin{tikzpicture}[>=Stealth, node distance=6mm and 0mm, every node/.style={inner sep=0,outer sep=0}]
    \node (sf1) {
      \begin{minipage}{0.95\columnwidth}
        \centering
        \includegraphics[width=\linewidth]{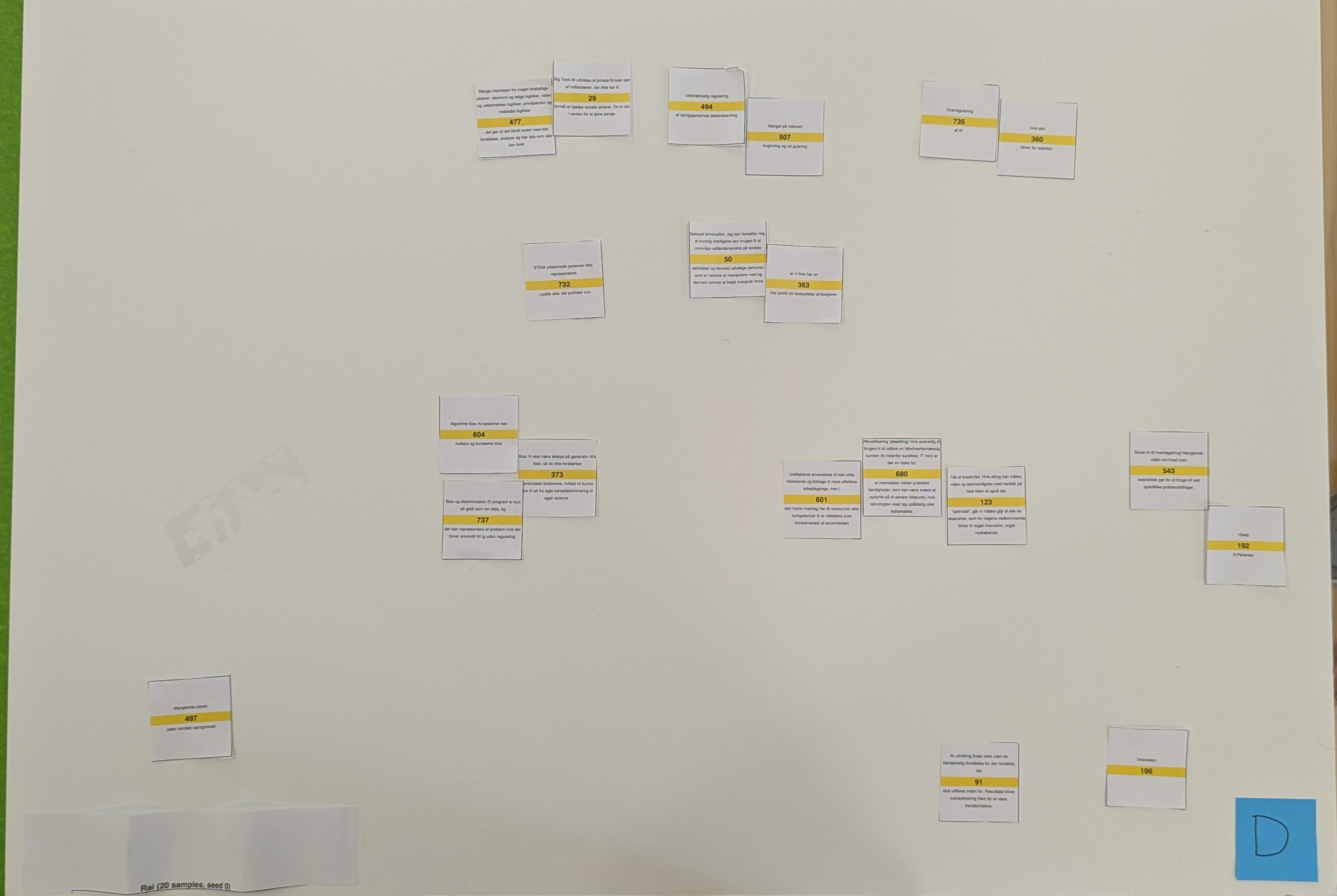}
        \subcaption{Raw image.}
        \label{fig:pipeline-raw-image}
      \end{minipage}
    };

    \node (sf2) [below=of sf1] {
      \begin{minipage}{0.95\columnwidth}
        \centering
        \includegraphics[width=\linewidth]{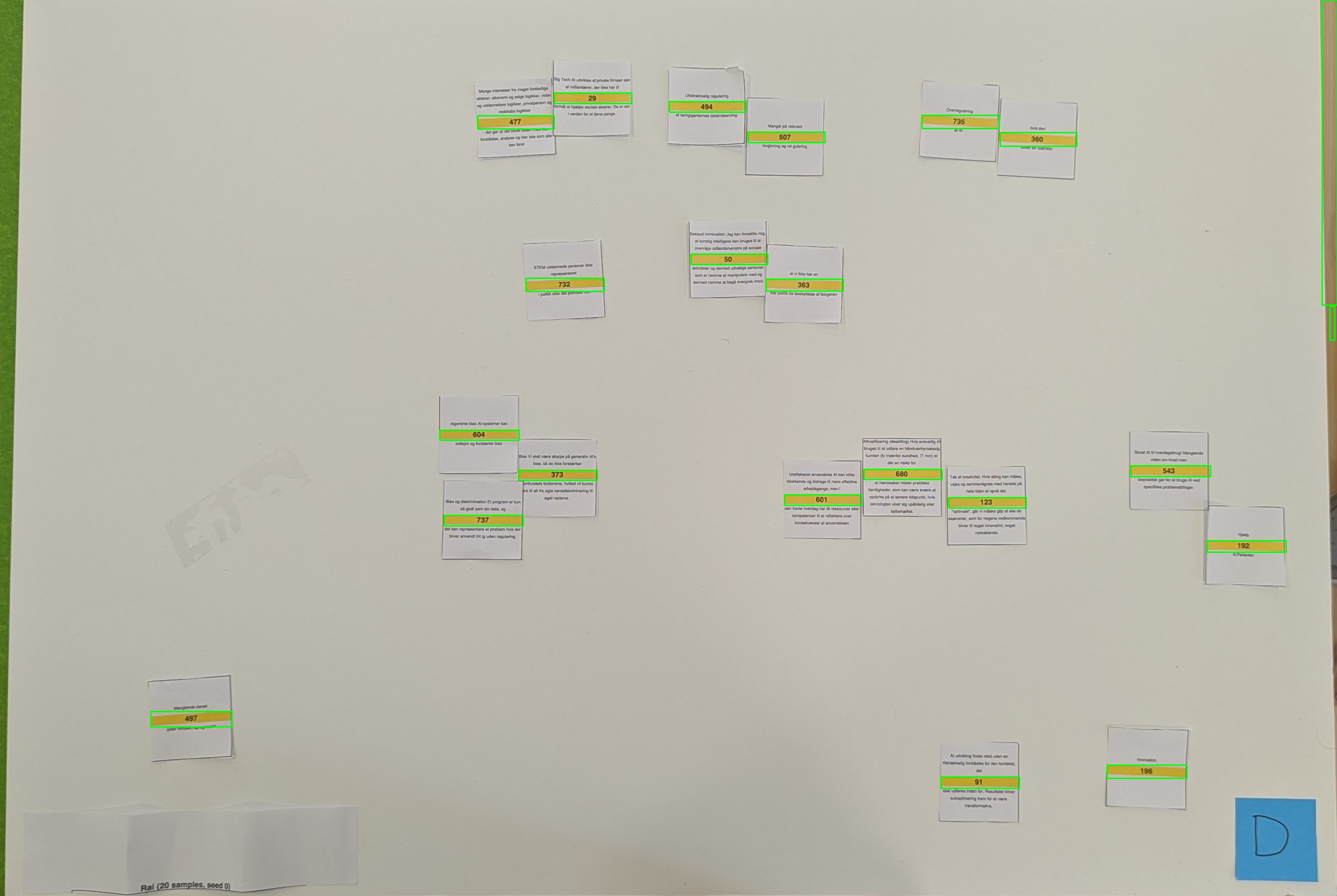}
        \subcaption{Bounding boxes (manual rectification of artefacts).}
        \label{fig:pipeline-bounding}
      \end{minipage}
    };

    \node (sf3) [below=of sf2] {
      \begin{minipage}{0.95\columnwidth}
        \centering
        \includegraphics[width=\linewidth]{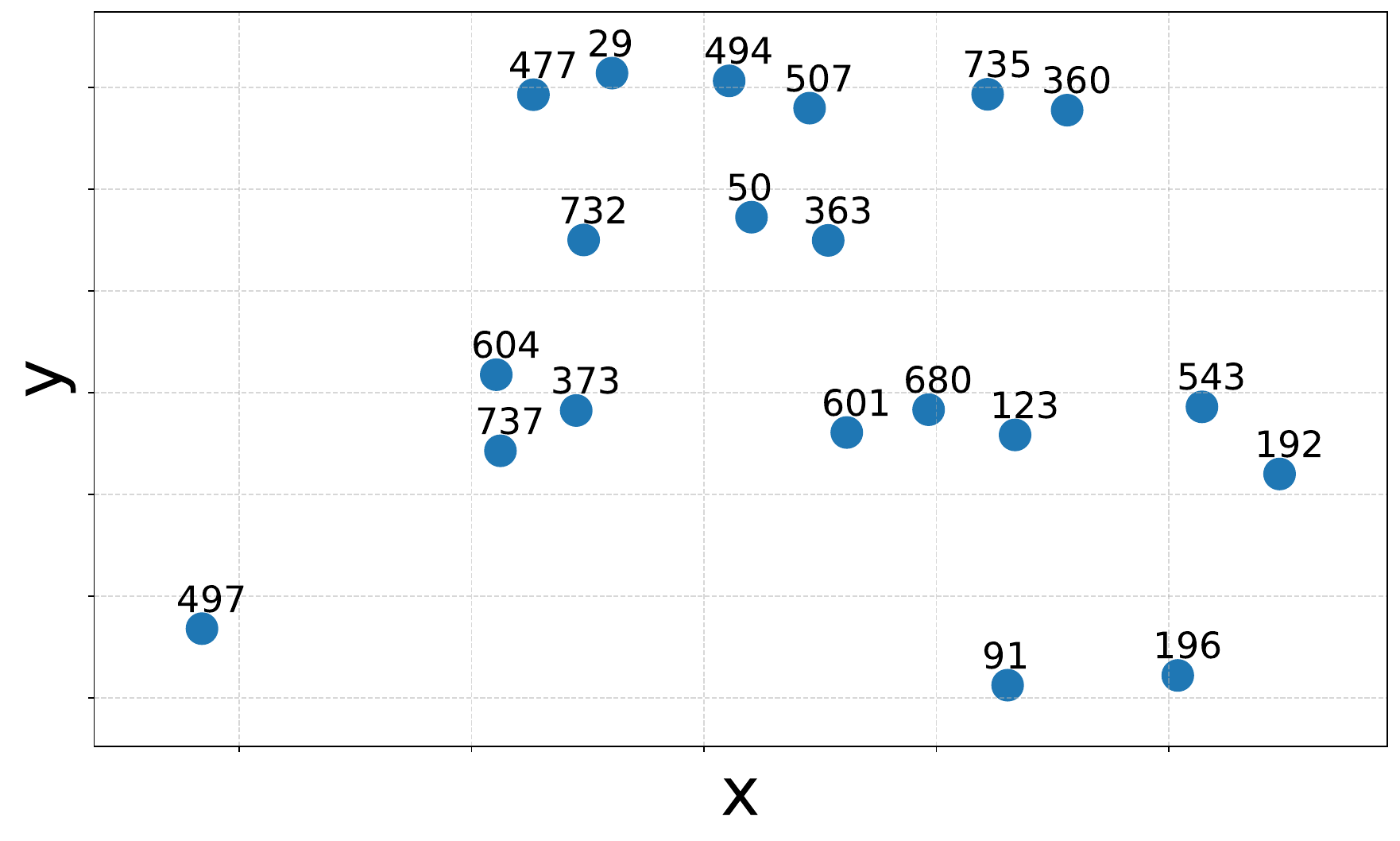}
        \subcaption{Extracted coordinates (optionally normalised).}
        \label{fig:pipeline-coordinates}
      \end{minipage}
    };

    \draw[->, thick] ($(sf1.south)-(0,1mm)$) -- ($(sf2.north)+(0,1mm)$);
    \draw[->, thick] ($(sf2.south)-(0,1mm)$) -- ($(sf3.north)+(0,1mm)$);
  \end{tikzpicture}

  \caption{Pipeline from raw images to coordinates. Top: raw image. Middle: constructed bounding boxes (artefacts such as empty boxes are manually removed). Bottom: final coordinates (can be normalised using image dimensions).}
  \label{fig:pipeline-overview}
\end{figure}

Physical-canvas arrangements were digitised using OpenCV \citep{bradskiOpencvLibrary2000} for canvas/ID detection and TrOCR \citep{liTrOCRTransformerbasedOptical2023} for ID recognition, followed by removing spurious detections and correcting misread IDs; no coordinates were modified. The web-app produces coordinates directly. 

Here, we illustrate how we go from raw images (Fig. \ref{fig:pipeline-raw-image}) to coordinates (Fig. \ref{fig:pipeline-coordinates}) using the physical pipeline.
The first step is to construct bounding boxes (Fig. \ref{fig:pipeline-bounding}). As the image shows, there can be artefacts (empty green boxes on the far right). These are the types of error that are manually rectified.
Once the bounding boxes and IDs have been extracted, we transform it into a csv file with coordinates (Fig. \ref{fig:pipeline-coordinates}). For some analyses, these are further normalised using the dimensions of the image.

In the web-app version of the pipeline, the coordinates are produced directly from the exercise.

\section{Participants} \label{app:participants}

\begin{table*}[t]
\centering
\small
\caption{Participant demographics (Policy).}
\label{tab:demographics}
\begin{tabular}{lcccccc}
\toprule
 & \textbf{P1} & \textbf{P2} & \textbf{P3} & \textbf{P4} & \textbf{P5} & \textbf{P6} \\
\midrule
Age              & 22-27  & 27-35  & 22-27  & 22-27  & 22-27  & 22-27  \\
Gender           & M  & F  & F  & M  & M  & M  \\
Education        & MSc & PhD & MSc & MSc & MSc & PhD \\
Field of study   & Design  & Anthropology  & Media  & Political Science  &  Computer Science & HCI  \\
Affiliation      & Partner & Partner & Partner & Partner & Partner & Ext. \\
\bottomrule
\end{tabular}
\end{table*}

Table \ref{tab:demographics} describes the demographics of our participants in the Policy. All participants were Danish. `Partner' denotes that the participants were recruited from our partner organisation (anonymised for review). `Ext' denotes external researcher.

For the Gov-AI exercise, Table \ref{tab:demographics-govai} describes the participants. All participants were PhD students recruited from the same university (anonymised for review).

\begin{table*}[t]
\centering
\small
\caption{Participant demographics (Gov-AI).}
\label{tab:demographics-govai}
\begin{tabular}{lcccccc}
\toprule
 & \textbf{P1} & \textbf{P2} & \textbf{P3} & \textbf{P4} & \textbf{P5} & \textbf{P6} \\
\midrule
Age              & 22-27  & 22-27  & 27-35  & 27-35  & 27-35  & 27-35  \\
Gender           & M  & M  & F  & M  & M  & M  \\
Education        & MSc & MSc & MSc & MSc & MSc & MSc \\
Field of study   & AI Fairness  & Privacy/AI  & AI Economics  & AI Economics  &  AI Evaluation  & AI Fairness  \\
Nationality      & Denmark & Netherlands & USA & USA & Germany & India \\
\bottomrule
\end{tabular}
\end{table*}

\section{Context drift} 
\label{app:context-drift}
\begin{figure}
    \centering
    \includegraphics[width=1.0\linewidth]{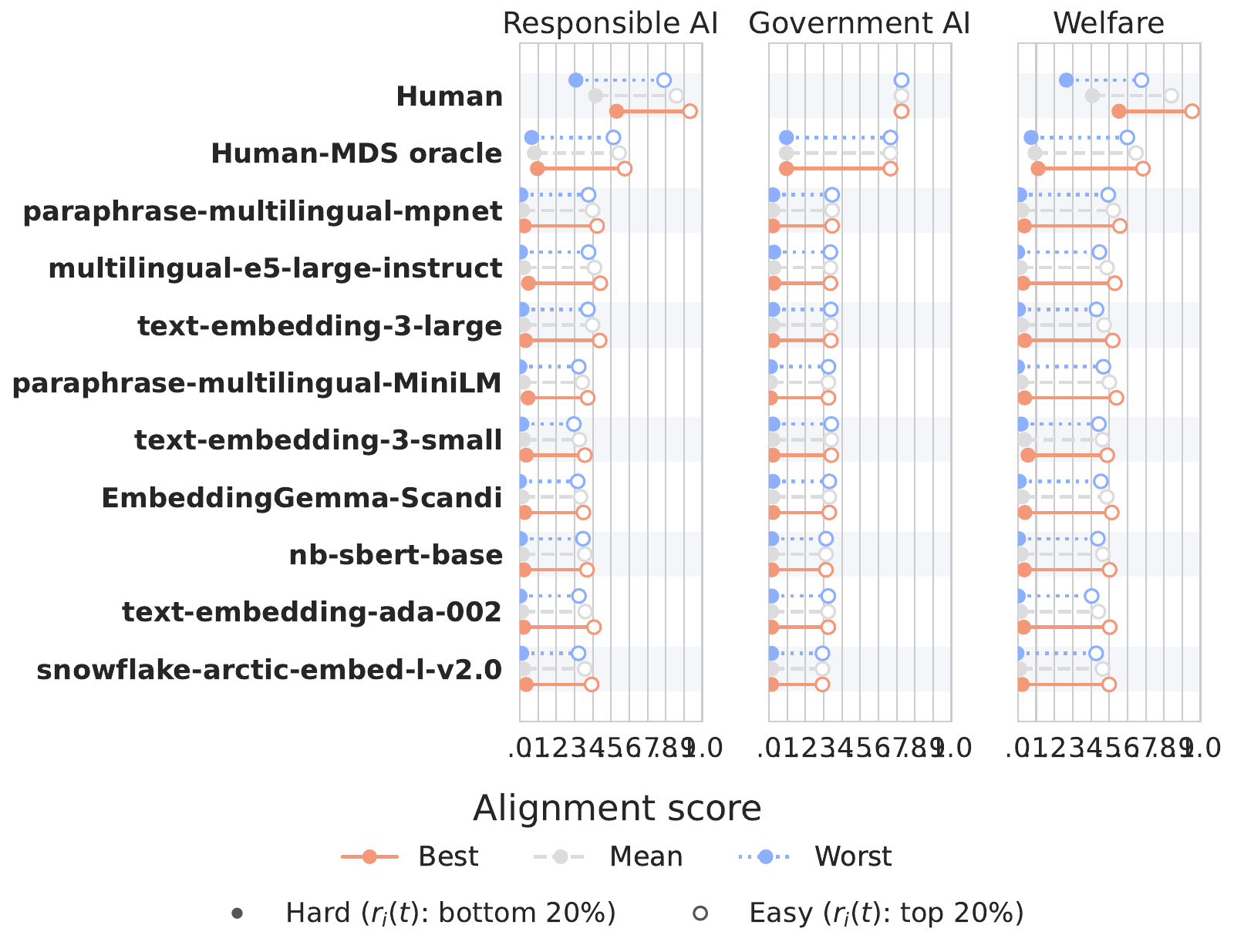}
    \caption{\textbf{Difficulty differences}: Comparison of scores of the most difficult ($r_i(t)$ in bottom quintile) versus easiest ($r_i(t)$  in top quintile) triplets. We observe a large gap in scores between human and top models for all levels.}
    \label{fig:difficulty}
\end{figure}

\begin{figure}
    \centering
    \includegraphics[width=\linewidth]{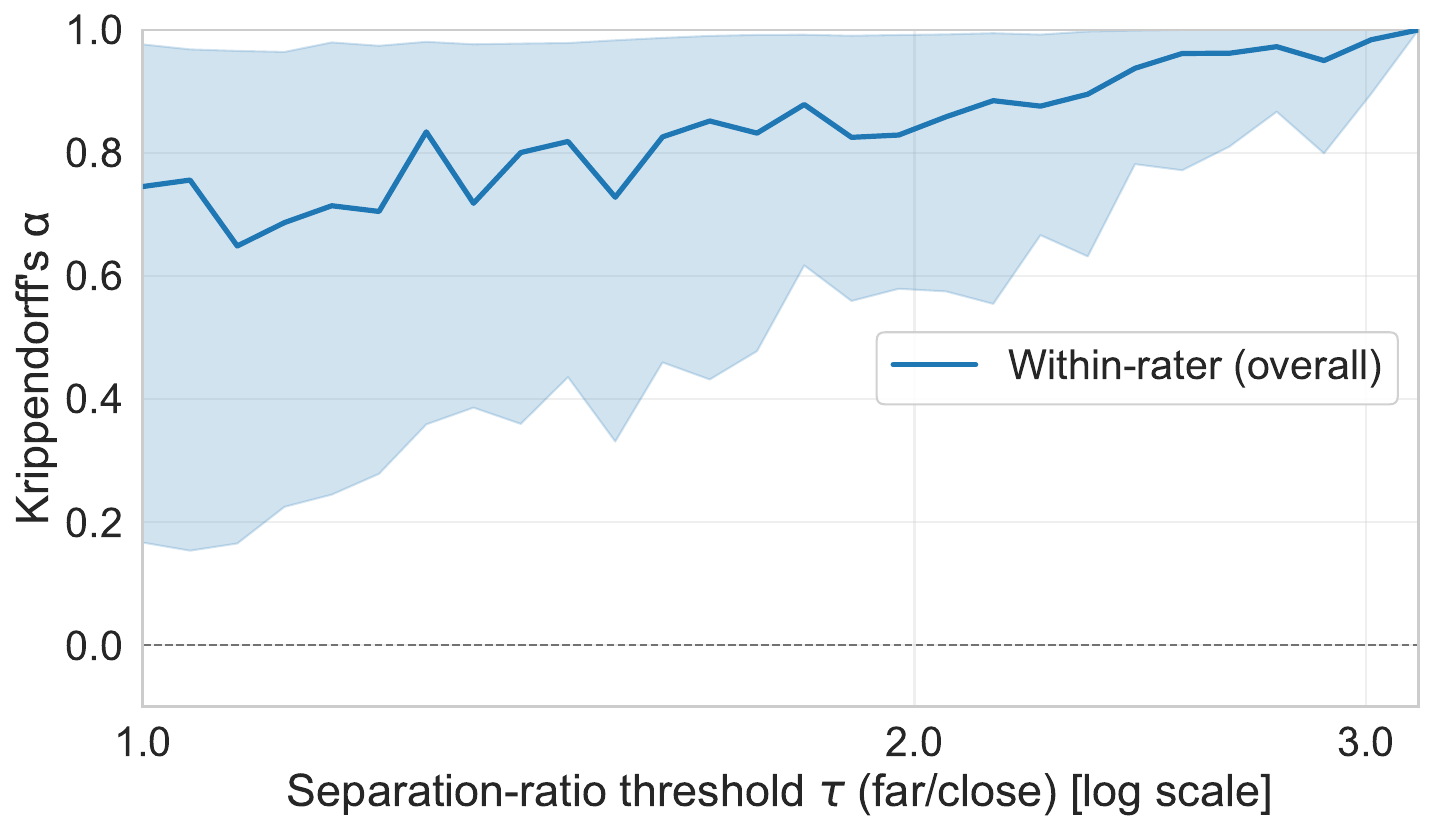}
    \caption{\textbf{Within-rater reliability}. The within-rater reliability is consistently high and increasing for higher $\tau$, similarly to between-rater reliability (Fig. \ref{fig:rq1-human-reliability})}
    \label{fig:within-rater}
\end{figure}

\begin{figure}
    \centering
    \includegraphics[width=\linewidth]{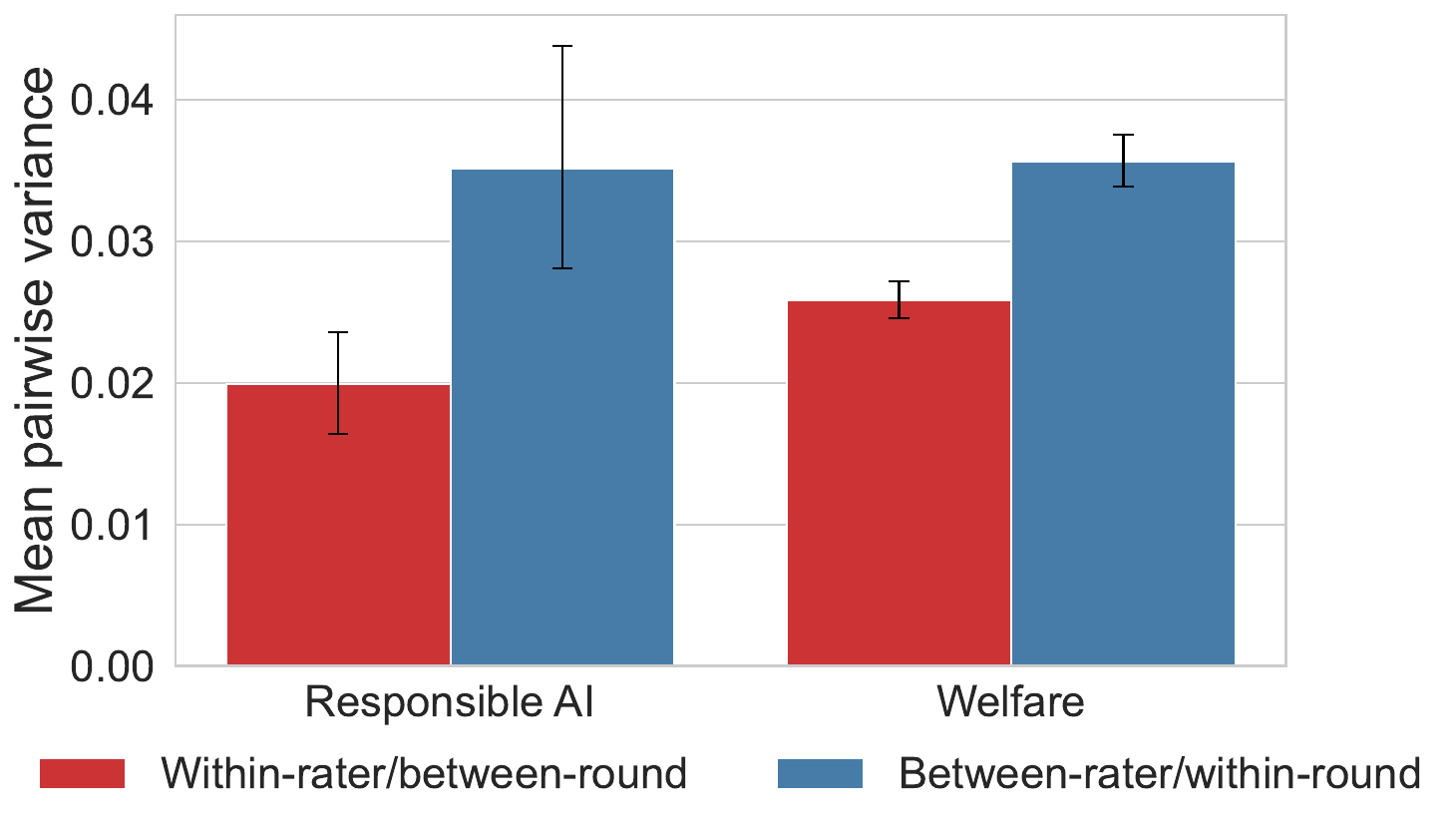}
    \caption{\textbf{Context-drift relative to rater noise}. Context-drift (operationalised as within-rater, between-round pairwise-distance variance) relative to variance between raters within a round. Context-drift is generally lower than noise, indicating no substantial influence on the results.}
    \label{fig:context-drift}
\end{figure}

Here, we evaluate potential \emph{context drift}, i.e., whether the relative positions of statements are stable across rounds for a given participant. Since the Stakeholder Grounding Exercise is inherently contextual, context drift might make it challenging to produce stable comparisons with embedding models.

We measure context drift in two ways: using within-rater reliability and measuring the variance of pairwise distances across rounds. The motivation for within-rater reliability is that it captures whether participants produce stable triplets across rounds, providing a local reliability measure. The pairwise distances, on the other hand, measure whether pairs have similar distances across rounds, providing a more global measure. For the pairwise, we compare against variation in pairwise distance between participants within a round, which provides a measure of the ``noise'' in a given round. 

We find consistently high within-rater reliability (Fig. \ref{fig:within-rater}). We also find that the pairwise context-drift is generally smaller than the between-rater noise (Fig. \ref{fig:context-drift}). Since our reliability analysis already absorbs rater variability into $\alpha(\tau)$, drift cannot be a dominant validity threat: it is at most as large as the noise $\alpha(\tau)$ already accounts for.

\section{Cluster ablations}
\label{app:kmeans}

\input{figs/cluster_consistency_table_adjusted_rand_index}

To assess robustness to clustering algorithm choice, we repeated the downstream analysis using k-means instead of Ward’s agglomerative clustering. The number of clusters was set to the average number of human clusters per round, consistent with the main analysis.

Model rankings under k-means were highly correlated with those obtained using Ward linkage (Spearman $\rho = 0.962$, $p = 2.21 \times 10^{-18}$).

The human--model performance gap in Adjusted Rand Index (ARI) was $0.194$ under Ward and $0.202$ under k-means ($\Delta = 0.009$), indicating that the magnitude of the gap is not sensitive to the clustering algorithm.

These results suggest that the observed downstream misalignment is not an artefact of the specific clustering procedure.

\paragraph{Sensitivity to the choice of \(K\).}
In the main clustering analysis, model clusters use the average number of clusters produced by the two human raters for the corresponding round. To test whether the downstream results depend on this human-derived \(K\), we repeated the analysis with internal model-side \(K\)-selection using silhouette score over \(K \in \{2,\ldots,10\}\), independently for each model and round.

Model rankings under silhouette-selected \(K\) were highly correlated with the main human-\(K\) rankings (Spearman \(\rho = 0.91\)). The correlation between stakeholder-grounding AUC and clustering ARI remained similar (\(0.79\) vs \(\GroundingClusteringCorrelation\)  for Policy in the main analysis); the corresponding correlation between MMTEB and clustering ARI was \MMTEBClusteringCorrelationAverage{} (averaged across Policy and Gov-AI). Thus, the conclusion that stakeholder grounding better predicts downstream clustering performance than MMTEB is not driven by the specific choice of \(K\). See Table \ref{tab:k-sensitivity}.

\input{figs/k_sensitivity_table}

\section{Metric and threshold robustness} \label{app:metric-robustness}

\paragraph{Normalised AUC over \(\alpha(\tau)\).}
For the main reliability and model-grounding results, we evaluate \(\alpha(\tau)\) on a log-spaced grid of separation-ratio thresholds and compute a normalised AUC in log-\(\tau\) space:

\begin{equation}
\begin{aligned}
\mathrm{AUC}_{\log \tau}
&= \frac{1}{\log \tau_{\max}-\log \tau_{\min}} \\
&\quad \times
\int_{\log \tau_{\min}}^{\log \tau_{\max}}
\alpha(\tau)\,\tau\log\tau .
\end{aligned}
\end{equation}

In implementation, we approximate this integral with the trapezoidal rule over the evaluated grid. The normalisation makes scores comparable across separation-threshold ranges.

\paragraph{Threshold and integration sensitivity.}
To test whether model rankings depend on the threshold grid, we re-ran the model-grounding leaderboard under alternative \(\tau_{\max}\), grid-size, and integration choices. Table~\ref{tab:auc-sensitivity} reports rank correlations with the main configuration. Rankings are stable across these choices, indicating that the main conclusions are not an artefact of the AUC parameterisation.

\input{figs/auc_sensitivity_table}

\paragraph{Expected disagreement.}
The main analysis uses the binary nominal form of Krippendorff's \(\alpha\). For each retained triplet, the outcome is one of two labels: the first comparison statement is closer to the anchor, or the second comparison statement is closer. Under random labelling, the probability of disagreement is therefore \(D_e = 0.5\), giving:
\begin{equation}
\alpha(\tau) = 1 - \frac{D_o(\tau)}{0.5}
          = 2P(\mathrm{agreement}) - 1 .
\end{equation}

\begin{figure}[t]
    \centering
    \includegraphics[width=\linewidth]{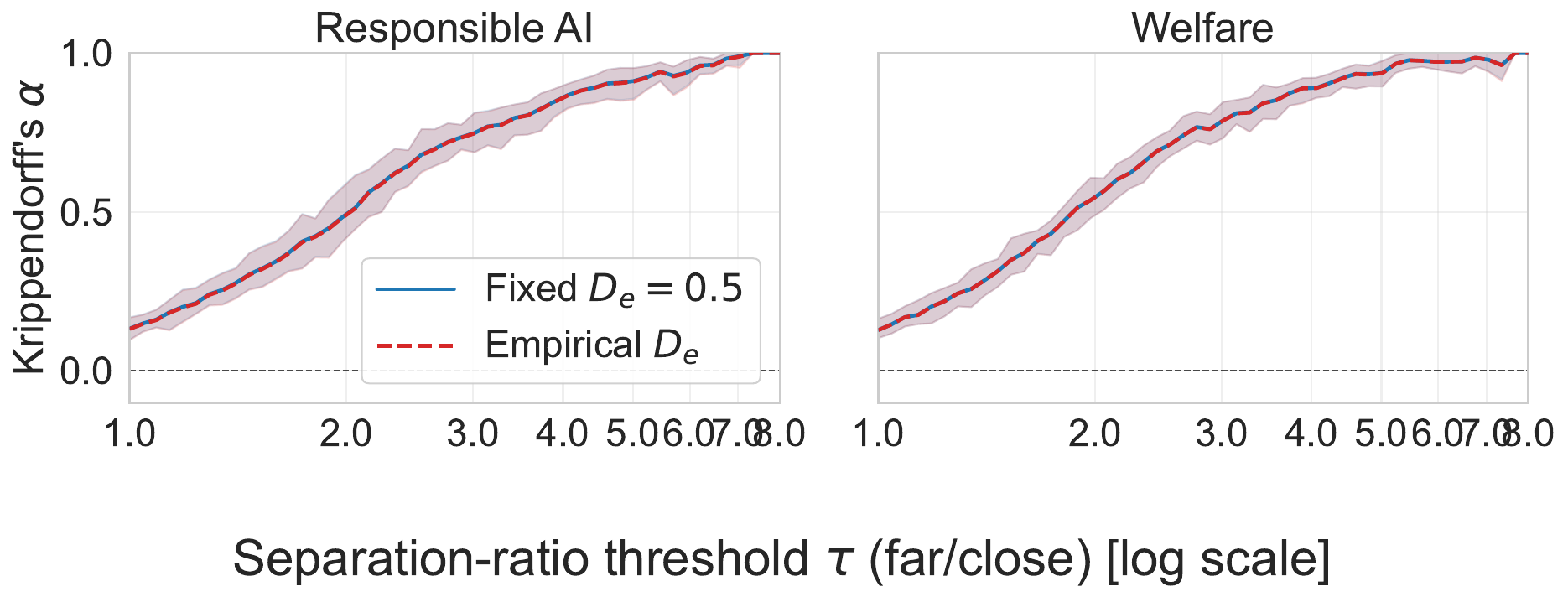}
    \caption{\textbf{Sensitivity to expected disagreement.} Comparison of the main binary-\(D_e=0.5\) \(\alpha(\tau)\) curves with curves using empirical marginal estimates of \(D_e\).}
    \label{fig:empirical-de-alpha}
\end{figure}

As a robustness check, we also compute \(D_e\) empirically from the observed marginal frequencies of the two triplet outcomes within each filtered dataset--group--threshold cell. Because the separation-ratio filter can slightly alter the balance of the two outcomes, this empirical \(D_e\) varies across \(\tau\). Fig.~\ref{fig:empirical-de-alpha} compares the resulting \(\alpha(\tau)\) curves with the main binary-\(D_e\) curves.

Finally, we show that the results are robust for different sets and permutations of raters (Fig. \ref{fig:rater-stability}). We describe the methodology for this in the main text (\secref{sec:rq2-methods}). Code is in our repository.\footnote{\url{https://github.com/jhrystrom/human-groundingscripts/rater_stability.py}}

\begin{figure}
    \centering
    \includegraphics[width=\linewidth]{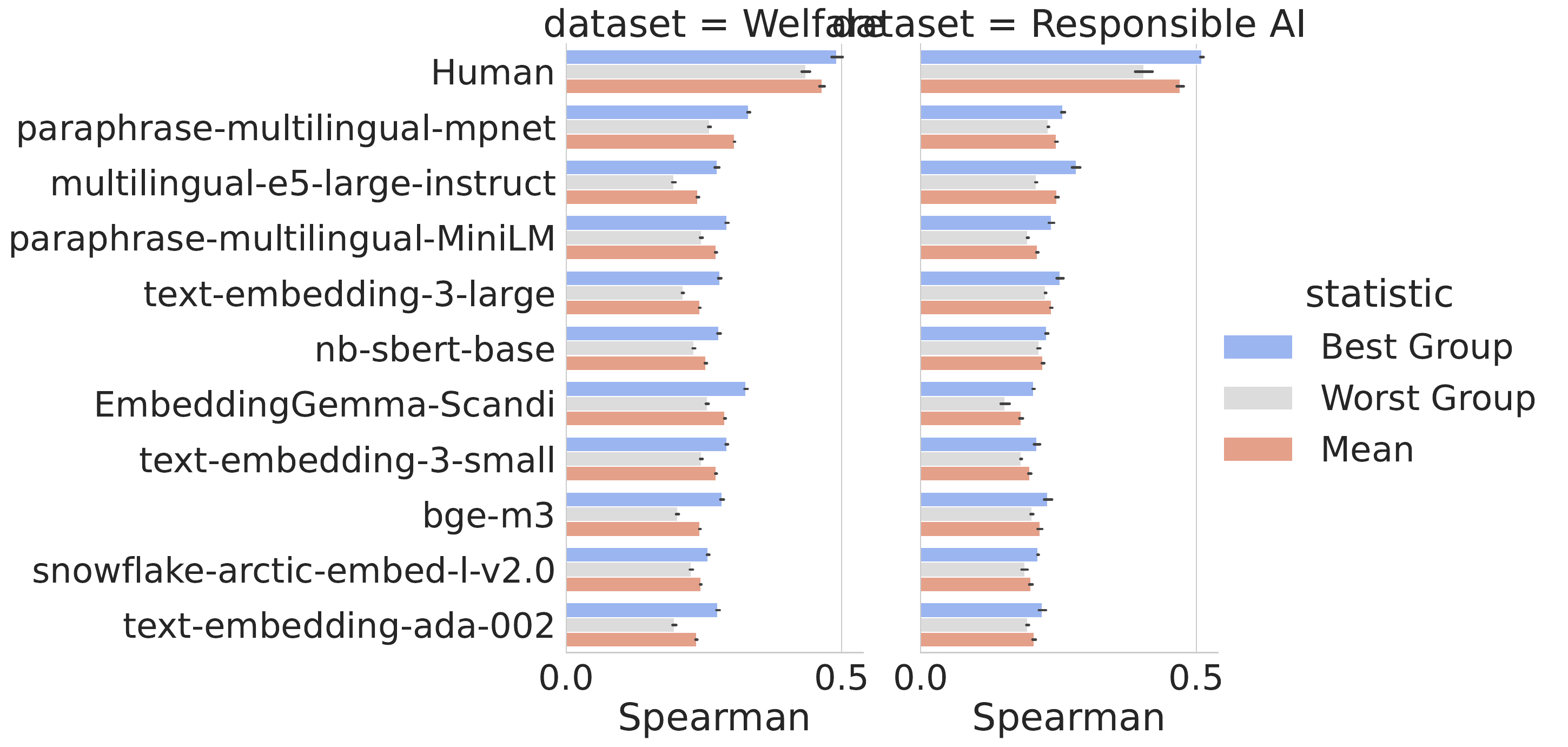}
    \caption{\textbf{Spearman metric}. We show qualitatively similar results while replacing our AUC-based alignment metric with a raw Spearman correlation of pairwise distances.}
    \label{fig:rq2-spearman-ablation}
\end{figure}

\paragraph{Spearman robustness.}
To illustrate the robustness of our results, we re-run the analysis with an alternative \citet{spearmanProofMeasurementAssociation1904} correlation of pairwise distances. For these metrics, we first extract all possible pairs across rounds. We then calculate the Spearman correlation for both inter-human participants and human-text embedding pairs. The results are shown in Fig. \ref{fig:rq2-spearman-ablation}. 

The results are qualitatively similar to the main metric. Humans still have a substantial gap to the best embedding model. We also find similar models on top such as paraphrase-multilingual-mpnet \citep{reimersSentenceBERTSentenceEmbeddings2019} and multilingual-e5-large-instruct \citep{wangMultilingualE5Text2024}. 

\subsection{Oracle construction} \label{app:oracle}
Each participant layout is centred and rescaled by its median pairwise distance to make arrangements comparable across rounds. We then construct a consensus dissimilarity matrix by averaging within-round Euclidean distances over all rounds in which a statement pair co-occurs \citep{borgModernMultidimensionalScaling2007}. We back-fill unobserved pairs with the mean observed dissimilarity so that MDS operates on a complete matrix; we only score observed triplets. We fit metric MDS \citep[SMACOF, 4 random initialisations, normalised Stress-1;][]{deleeuwApplicationsConvexAnalysis2005} at candidate dimensionalities $q \in {2, 5, 10, 20, 30, 50, 75, 100}$ and select the smallest $q$ beyond which stress improves by less than 1\% \citep{borgModernMultidimensionalScaling2007}. Because the oracle is fitted to reproduce Euclidean distances, it is evaluated under Euclidean distance; every other step of the triplet pipeline is identical to the embedding models.

The sweep results in the following values of $q$ and the resulting stress values for the fit. For Responsible AI, we find $q=\ResponsibleAIOracleQ$ and a stress of \ResponsibleAIOracleStress. For Welfare, we find $q=\WelfareOracleQ$ and a stress of \WelfareOracleStress. Finally, for Government AI, we find $q=\GovAIOracleQ$ and a stress of \GovAIOracleStress.

\section{Model-model reliability}
\begin{figure}
    \centering
    \includegraphics[width=\linewidth]{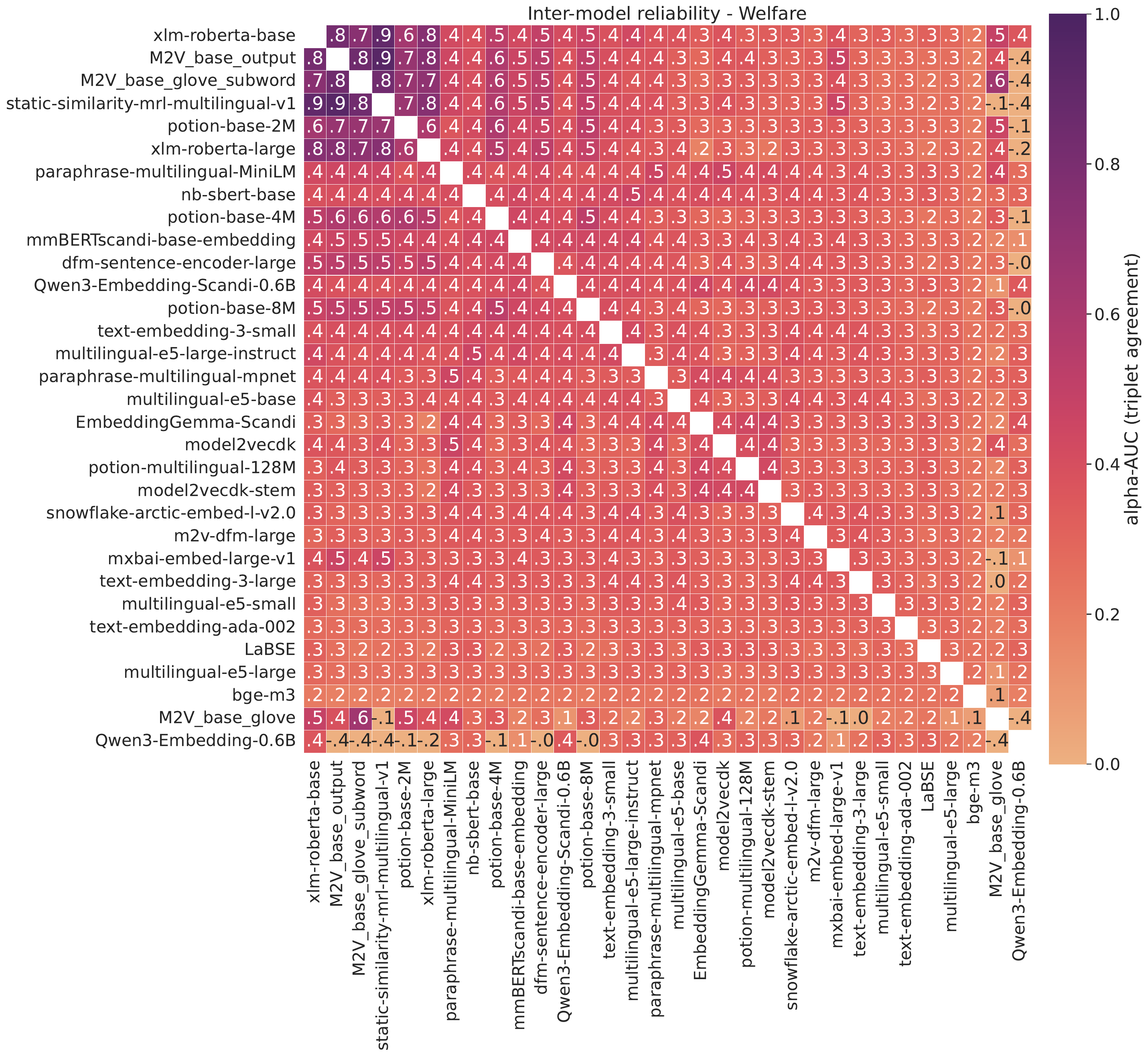}
    \caption{\textbf{Model-model reliability.} Embedding models have varying inter-model reliability from 0 to 0.9. Typically, they're in a similar range to model-human reliability (see Fig. \ref{fig:rq2-scores}).}
    \label{fig:intermodel}
\end{figure}

Fig. \ref{fig:intermodel} shows the inter-model reliability for all pairs of models. It shows large variations, albeit in a similar range to the human scores.

\section{Embedding models and lexical baselines} \label{app:embedding-list}
\input{figs/model_type_rank_delta_gov-ai_policy}
Below is a list of all the embedding models evaluated in this study. 
In addition to neural embedding models, we include two lexical baselines to test whether stakeholder-grounding performance can be explained by surface-form similarity alone. The first, \texttt{tfidf-char35}, uses a character \(3\)--\(5\)-gram TF--IDF representation with sublinear term frequency and \(\ell_2\)-normalisation. Character \(n\)-grams are used because the Policy datasets are primarily Danish while Gov-AI is English. The second, \texttt{jaccard-binary}, uses binary word-token features; under Euclidean distance this corresponds to a monotone transformation of token-set symmetric difference, providing a simple lexical-overlap baseline.

Table~\ref{tab:lexical-baselines} reports their alignment AUC and clustering ARI relative to the full model set.

\input{figs/lexical_baselines_table}

\begin{itemize}
  \item \href{https://huggingface.co/minishlab/potion-multilingual-128M}{potion-multilingual-128M} \citep{tulkensModel2VecFastStateoftheart2024}
  \item \href{https://huggingface.co/minishlab/potion-base-8M}{potion-base-8M} \citep{tulkensModel2VecFastStateoftheart2024}
  \item \href{https://huggingface.co/minishlab/potion-base-4M}{potion-base-4M} \citep{tulkensModel2VecFastStateoftheart2024}
  \item \href{https://huggingface.co/minishlab/potion-base-2M}{potion-base-2M} \citep{tulkensModel2VecFastStateoftheart2024}
  \item \href{https://huggingface.co/minishlab/M2V_base_glove}{M2V\_base\_glove} \citep{tulkensModel2VecFastStateoftheart2024}
  \item \href{https://huggingface.co/minishlab/M2V_base_glove_subword}{M2V\_base\_glove\_subword} \citep{tulkensModel2VecFastStateoftheart2024}
  \item \href{https://huggingface.co/minishlab/M2V_base_output}{M2V\_base\_output} \citep{tulkensModel2VecFastStateoftheart2024}
  \item \href{https://huggingface.co/rasgaard/m2v-dfm-large}{m2v-dfm-large} \citep{tulkensModel2VecFastStateoftheart2024,enevoldsenScandinavianEmbeddingBenchmarks2024}
  \item \href{https://huggingface.co/andersborges/model2vecdk}{model2vecdk} \citep{tulkensModel2VecFastStateoftheart2024}
  \item \href{https://huggingface.co/andersborges/model2vecdk-stem}{model2vecdk-stem} \citep{tulkensModel2VecFastStateoftheart2024}

  \item \href{https://huggingface.co/Qwen/Qwen3-Embedding-0.6B}{Qwen3-Embedding-0.6B} \citep{zhangQwen3EmbeddingAdvancing2025}
  \item \href{https://huggingface.co/intfloat/multilingual-e5-large-instruct}{multilingual-e5-large-instruct} \citep{wangMultilingualE5Text2024}
  \item \href{https://huggingface.co/emillykkejensen/EmbeddingGemma-Scandi-300m}{EmbeddingGemma-Scandi-300m} \citep{veraEmbeddingGemmaPowerfulLightweight2025}
  \item \href{https://huggingface.co/emillykkejensen/Qwen3-Embedding-Scandi-0.6B}{Qwen3-Embedding-Scandi-0.6B} \citep{zhangQwen3EmbeddingAdvancing2025}

  \item \href{https://huggingface.co/intfloat/multilingual-e5-large}{multilingual-e5-large} \citep{wangMultilingualE5Text2024}
  \item \href{https://huggingface.co/BAAI/bge-m3}{bge-m3} \citep{chenM3embeddingMultilingualityMultifunctionality2024}
  \item \href{https://huggingface.co/intfloat/multilingual-e5-base}{multilingual-e5-base} \citep{wangMultilingualE5Text2024}
  \item \href{https://huggingface.co/Snowflake/snowflake-arctic-embed-l-v2.0}{snowflake-arctic-embed-l-v2.0} \citep{yuArcticembed20Multilingual2024}
  \item \href{https://huggingface.co/intfloat/multilingual-e5-small}{multilingual-e5-small} \citep{wangMultilingualE5Text2024}
  \item \href{https://huggingface.co/NbAiLab/nb-sbert-base}{nb-sbert-base} \citep{reimersSentenceBERTSentenceEmbeddings2019}
  \item \href{https://huggingface.co/KennethEnevoldsen/dfm-sentence-encoder-large}{dfm-sentence-encoder-large} \citep{enevoldsenScandinavianEmbeddingBenchmarks2024}
  \item \href{https://huggingface.co/emillykkejensen/Qwen3-Embedding-Scandi-0.6B}{mmBERTscandi-base-embedding} \citep{zhangQwen3EmbeddingAdvancing2025}
  \item \href{https://huggingface.co/sentence-transformers/LaBSE}{LaBSE} \citep{fengLanguageagnosticBERTSentence2022}
  \item \href{https://huggingface.co/sentence-transformers/paraphrase-multilingual-mpnet-base-v2}{paraphrase-multilingual-mpnet-base-v2} \citep{reimersSentenceBERTSentenceEmbeddings2019}
  \item \href{https://huggingface.co/FacebookAI/xlm-roberta-large}{xlm-roberta-large} \citep{conneauUnsupervisedCrosslingualRepresentation2020}
  \item \href{https://huggingface.co/sentence-transformers/paraphrase-multilingual-MiniLM-L12-v2}{paraphrase-multilingual-MiniLM-L12-v2} \citep{reimersSentenceBERTSentenceEmbeddings2019}
  \item \href{https://huggingface.co/FacebookAI/xlm-roberta-base}{xlm-roberta-base} \citep{conneauUnsupervisedCrosslingualRepresentation2020}
  \item \href{https://huggingface.co/sentence-transformers/static-similarity-mrl-multilingual-v1}{static-similarity-mrl-multilingual-v1} \citep{reimersSentenceBERTSentenceEmbeddings2019}
  \item \href{https://huggingface.co/mixedbread-ai/mxbai-embed-large-v1}{mxbai-embed-large-v1} \citep{liAoEAngleoptimizedEmbeddings2024}

  \item \href{https://platform.openai.com/docs/guides/embeddings}{text-embedding-3-large}
  \item \href{https://platform.openai.com/docs/guides/embeddings}{text-embedding-ada-002} 
  \item \href{https://platform.openai.com/docs/guides/embeddings}{text-embedding-3-small}
  \item \texttt{tfidf-char35}: character \(3\)--\(5\)-gram TF--IDF lexical baseline.
  \item \texttt{jaccard-binary}: binary word-token lexical-overlap baseline.
\end{itemize}

\section{Instruction prompts} \label{app:instruction}
For instruction-tuned embedding models, we used the developer-recommended generic similarity instructions and beyond the variants below, we did not optimise prompts.

The following prompt templates were applied via the \texttt{prompt=} argument in \texttt{SentenceTransformer.encode()} \citep{reimersSentenceBERTSentenceEmbeddings2019}:

\begin{itemize}
    \item \textbf{multilingual-e5-large-instruct:} \citep{wangMultilingualE5Text2024}\\
    \texttt{"Instruct: Cluster similar statements\textbackslash nQuery: "}

    \item \textbf{EmbeddingGemma-Scandi-300m:} \citep{veraEmbeddingGemmaPowerfulLightweight2025} \\
    \texttt{"task: sentence similarity | query: "}
\end{itemize}

All other embedding models were used without additional prompting.

\begin{figure}
    \centering
    \includegraphics[width=\linewidth]{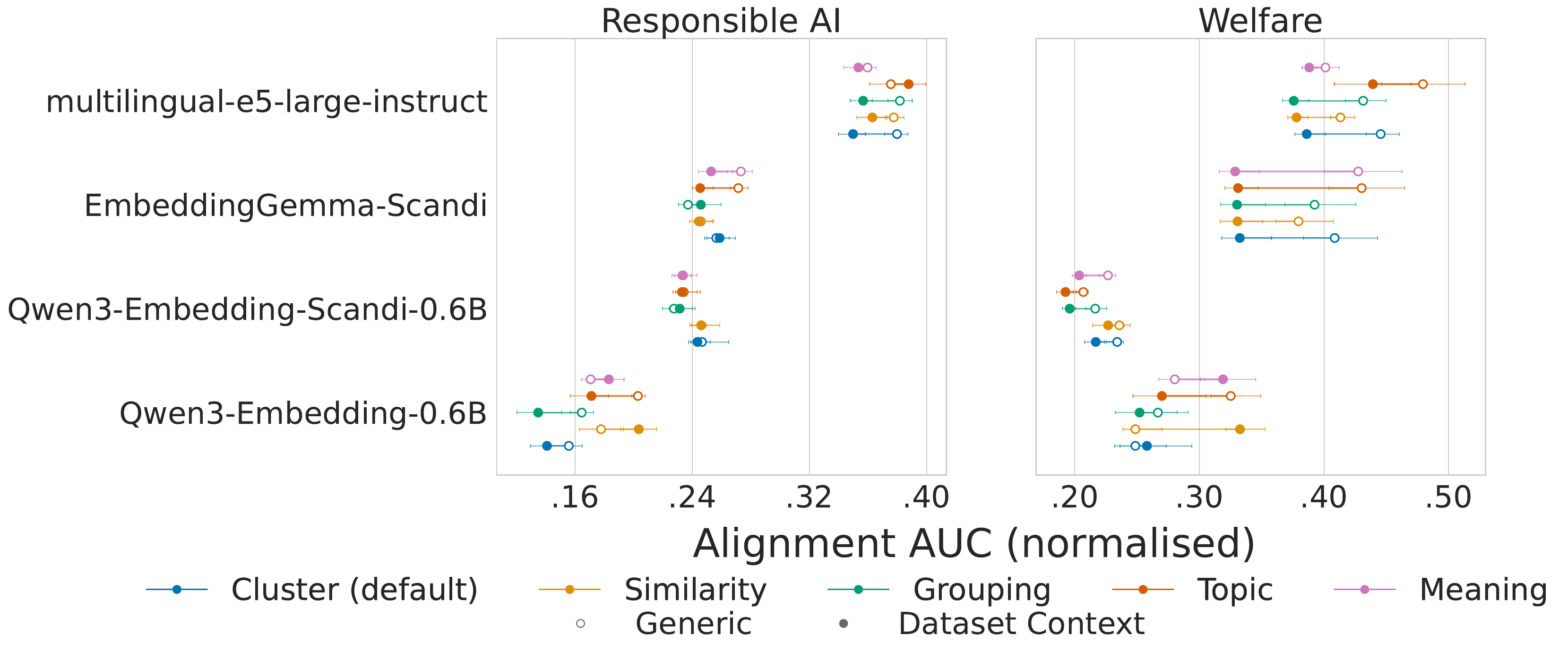}
    \caption{\textbf{Variation across instruction formats.} Prompt variations and context-specific instructions have varying effects.}
    \label{fig:instruct-results}
\end{figure}

We experimented with prepending topic-specific prompts to the instructions in the format, \texttt{"You are embedding policy problem statements in the domain of \{DOMAIN\}."}, where DOMAIN was either ``responsible AI'', ``the future of Danish municipal welfare'', or ``Government AI applications''. As shown in Fig. \ref{fig:instruct-results}, we found some variation in performance across the conditions. As suggested by \citet{kostiukOnePromptNot2026}, we additionally experimented with five variations of the prompt resulting in 10 total variations per model. The results can be seen in Fig. \ref{fig:instruct-results}. We find a maximum gain of \MaximumInstructionGain{}, which is consistent with \citet{suOneEmbedderAny2023}. 

A full list of prompts can be found in the repository \url{https://github.com/jhrystrom/human-grounding}

\section{Respondent demographics} \label{app:respondent-demo}
\input{figs/respondent_demographics}
Table \ref{tab:sample-composition} shows the composition of the respondents in the two datasets. Overall, both datasets are relatively balanced, with a slight majority of women in the Responsible AI dataset. Since Gov-AI is institutional, no demographic information is available. 

\paragraph{Group-wise triplet counts.}
Because the separation-ratio filter changes the retained triplet set as \(\tau\) increases, group-wise sample sizes vary across thresholds. Table~\ref{tab:group-triplet-counts} reports retained triplet counts for representative thresholds.

\input{figs/fairness_triplet_counts}

\paragraph{Rater-pair coverage.} Between-rater reliability is computed over pairs of participants who co-rated
at least one round. Of the $\binom{\MeanPanelRaters}{2} =
\MeanPossibleRaterPairs$ possible pairs per panel, on average
\MeanContributingRaterPairs{} contributed at least one retained triplet, with a
median of \PooledMedianCoRatedRounds{} co-rated rounds per contributing pair.
Pairs that never share a round contribute nothing to $\mathcal{T}_{i,j}$. We
show per-panel statistics in Table~\ref{tab:coverage}.

\paragraph{Bootstrap tests for group-wise gaps.}
For each model and dataset, we compute the best--worst group gap as
\[
\Delta_{\mathrm{group}} =
\alpha_{\mathrm{AUC}}^{\max}
-
\alpha_{\mathrm{AUC}}^{\min},
\]
where AUCs are drawn from hierarchical-bootstrap replicates across author groups defined by the source (anchor) statement of each triplet. 
We report results for the highest-mean-AUC model per dataset. To isolate group effects from surface-level text differences, we also compute an adjusted gap $\Delta_{\mathrm{adj}}$: the max-minus-min in predicted $P(\text{embedding correct})$ from a logistic regression of triplet correctness on group dummies plus controls---$\log$ token counts of the source, closer, and farther statements and Jaccard token overlap between source--closer and source--farther pairs. Adjusted predictions are evaluated at the in-sample mean of the controls; each bootstrap replicate resamples raters with replacement and refits the regression. Bootstrap $p$-values are one-sided proportions of replicates with $\Delta \le 0$. Table~\ref{tab:group-gap-controlled} reports the resulting gaps, confidence intervals, and $p$-values for both the unadjusted and adjusted analyses.

\input{figs/fairness_group_gap_controlled}
\input{figs/statement_coverage_table}

\section{Computational resources \& Licenses} \label{app:resource-license}
All experiments were run on a single H100 GPU. However, given the relatively small size of the embedding models ($<1B$ parameters; \ref{app:embedding-list}) it could be replicated on even smaller hardware. The total time for running all analyses is around 1 hour. Replication code and data are provided here: \url{https://github.com/jhrystrom/human-grounding}. For full reproduction, simply run \texttt{just reproduce-full} (once data has been downloaded).

We release the code and associated data under an MIT license. For model licenses see links in Appendix \ref{app:embedding-list}. The Federal Agency AI Use Case Inventory data we use is a work of the U.S. federal government and is in the public domain.

\input{latex/appendix_triplets}

\end{document}

%% file: latex/key-values.tex
\newcommand{\MeanPanelRaters}{6}

\newcommand{\MeanPossibleRaterPairs}{15}

\newcommand{\MeanContributingRaterPairs}{8.3}

\newcommand{\PooledMedianCoRatedRounds}{2}

\newcommand{\ResponsibleAIMenAUC}{0.64}

\newcommand{\ResponsibleAIWomenAUC}{0.65}

\newcommand{\ResponsibleAIMenAUCLO}{0.63}

\newcommand{\ResponsibleAIMenAUCHI}{0.65}

\newcommand{\ResponsibleAIWomenAUCLO}{0.64}

\newcommand{\ResponsibleAIWomenAUCHI}{0.66}

\newcommand{\BestEmbeddingModel}{paraphrase-multilingual-mpnet}

\newcommand{\SecondEmbeddingModel}{multilingual-e5-large-instruct}

\newcommand{\BestEmbeddingModelAUC}{0.39}

\newcommand{\ResponsibleAIModelGap}{26.9}

\newcommand{\WelfareModelGap}{18.2}

\newcommand{\GovAIModelGap}{16.1}

\newcommand{\ResponsibleAIModelGapLO}{25.7}

\newcommand{\ResponsibleAIModelGapHI}{28.7}

\newcommand{\WelfareModelGapLO}{16.2}

\newcommand{\WelfareModelGapHI}{20.3}

\newcommand{\GovAIModelGapLO}{11.0}

\newcommand{\GovAIModelGapHI}{20.9}

\newcommand{\ResponsibleAIOracleGap}{14.7}

\newcommand{\WelfareOracleGap}{13.5}

\newcommand{\GovAIOracleGap}{29.9}

\newcommand{\ResponsibleAIOracleStress}{0.21}

\newcommand{\WelfareOracleStress}{0.21}

\newcommand{\GovAIOracleStress}{0.22}

\newcommand{\ResponsibleAIOracleQ}{20}

\newcommand{\WelfareOracleQ}{10}

\newcommand{\GovAIOracleQ}{10}

\newcommand{\MMTEBGroundedSpearman}{0.62}

\newcommand{\MMTEBGroundedPValue}{0.0001}

\newcommand{\ResponsibleAIGroupGap}{0.028}

\newcommand{\WelfareGroupGap}{0.068}

\newcommand{\ResponsibleAIGroupGapLO}{0.026}

\newcommand{\ResponsibleAIGroupGapHI}{0.029}

\newcommand{\WelfareGroupGapLO}{0.062}

\newcommand{\WelfareGroupGapHI}{0.073}

\newcommand{\PolicyHumanARI}{0.33}

\newcommand{\PolicyHumanARILO}{0.26}

\newcommand{\PolicyHumanARIHI}{0.41}

\newcommand{\GroundingClusteringCorrelation}{0.86}

\newcommand{\GroundingClusteringCorrelationLO}{0.62}

\newcommand{\GroundingClusteringCorrelationHI}{0.94}

\newcommand{\MMTEBClusteringCorrelation}{0.56}

\newcommand{\MMTEBClusteringCorrelationLO}{0.20}

\newcommand{\MMTEBClusteringCorrelationHI}{0.79}

\newcommand{\MMTEBClusteringCorrelationAverage}{0.57}

\newcommand{\MaximumInstructionGain}{0.080}

\newcommand{\HeadlineGapMinimum}{18}

\newcommand{\HeadlineGapMaximum}{27}

\newcommand{\OracleName}{Human-MDS oracle}

%% file: figs/cluster_consistency_table_adjusted_rand_index.tex
\begin{table*}[t]
\centering
\small
\caption{Clustering consistency with human participants (Adjusted Rand Index) for all models, ranked by Borda count across experiments.}\label{tab:cluster-consistency-adjusted-rand-index}
\begin{tabular}{rlcc}
\toprule
\textbf{Rank} & \textbf{Model} & \textbf{Policy} & \textbf{Gov-AI} \\
\midrule
1 & \emph{Human} & \textbf{0.33 [0.26; 0.41]} & 0.16 [0.11; 0.21] \\
2 & multilingual-e5-large-instruct & 0.27 [0.22; 0.33] & 0.19 [0.14; 0.24] \\
3 & EmbeddingGemma-Scandi & 0.26 [0.21; 0.31] & 0.19 [0.14; 0.24] \\
4 & text-embedding-3-large & 0.23 [0.18; 0.29] & 0.19 [0.13; 0.25] \\
5 & text-embedding-ada-002 & 0.18 [0.13; 0.25] & 0.19 [0.14; 0.25] \\
6 & bge-m3 & 0.20 [0.15; 0.24] & 0.17 [0.12; 0.23] \\
7 & snowflake-arctic-embed-l-v2.0 & 0.21 [0.16; 0.27] & 0.16 [0.10; 0.22] \\
8 & paraphrase-multilingual-mpnet & 0.20 [0.16; 0.25] & 0.14 [0.09; 0.18] \\
9 & text-embedding-3-small & 0.15 [0.11; 0.21] & \textbf{0.20 [0.15; 0.25]} \\
10 & nb-sbert-base & 0.18 [0.13; 0.23] & 0.16 [0.13; 0.20] \\
11 & tfidf-char35 & 0.22 [0.18; 0.26] & 0.10 [0.05; 0.16] \\
12 & LaBSE & 0.20 [0.15; 0.26] & 0.11 [0.07; 0.16] \\
13 & multilingual-e5-base & 0.15 [0.11; 0.20] & 0.17 [0.12; 0.23] \\
14 & Qwen3-Embedding-Scandi-0.6B & 0.16 [0.12; 0.19] & 0.17 [0.12; 0.24] \\
15 & multilingual-e5-large & 0.17 [0.12; 0.21] & 0.14 [0.10; 0.19] \\
16 & m2v-dfm-large & 0.19 [0.14; 0.24] & 0.11 [0.07; 0.14] \\
17 & mmBERTscandi-base-embedding & 0.13 [0.10; 0.17] & 0.17 [0.13; 0.22] \\
18 & paraphrase-multilingual-MiniLM & 0.16 [0.11; 0.21] & 0.13 [0.09; 0.18] \\
19 & potion-multilingual-128M & 0.19 [0.14; 0.24] & 0.10 [0.06; 0.15] \\
20 & model2vecdk & 0.18 [0.13; 0.22] & 0.08 [0.04; 0.13] \\
21 & model2vecdk-stem & 0.17 [0.13; 0.21] & 0.11 [0.07; 0.17] \\
22 & multilingual-e5-small & 0.15 [0.10; 0.19] & 0.12 [0.07; 0.17] \\
23 & dfm-sentence-encoder-large & 0.11 [0.06; 0.16] & 0.16 [0.10; 0.22] \\
24 & Qwen3-Embedding-0.6B & 0.11 [0.07; 0.16] & 0.15 [0.10; 0.20] \\
25 & mxbai-embed-large-v1 & 0.07 [0.04; 0.10] & 0.17 [0.12; 0.22] \\
26 & potion-base-8M & 0.05 [0.02; 0.09] & 0.13 [0.09; 0.18] \\
27 & potion-base-2M & 0.04 [0.01; 0.07] & 0.14 [0.09; 0.18] \\
28 & potion-base-4M & 0.04 [0.00; 0.07] & 0.12 [0.08; 0.15] \\
29 & xlm-roberta-base & 0.06 [0.03; 0.10] & 0.07 [0.04; 0.12] \\
30 & M2V\_base\_glove\_subword & 0.04 [0.01; 0.06] & 0.06 [0.03; 0.10] \\
31 & M2V\_base\_output & 0.03 [0.00; 0.05] & 0.08 [0.04; 0.14] \\
32 & static-similarity-mrl-multilingual-v1 & 0.04 [0.02; 0.06] & 0.04 [0.00; 0.08] \\
33 & xlm-roberta-large & 0.03 [0.00; 0.05] & 0.07 [0.04; 0.10] \\
34 & M2V\_base\_glove & 0.02 [-0.01; 0.05] & 0.06 [0.03; 0.10] \\
35 & jaccard-binary & 0.01 [-0.00; 0.03] & 0.03 [0.01; 0.06] \\
\bottomrule
\end{tabular}
\end{table*}

%% file: figs/k_sensitivity_table.tex
\begin{table*}[t]
\centering
\small
\caption{Sensitivity of downstream clustering results to the choice of $K$.}
\label{tab:k-sensitivity}
\begin{tabular}{lccc}
\toprule
\textbf{Clustering setup} & \textbf{ARI rank $\rho$} & \textbf{Grounding--ARI $\rho$} & \textbf{MMTEB--ARI $\rho$} \\
\midrule
Human-derived $K$ & 1.00 & 0.86 & 0.57 \\
Silhouette-selected $K$ & 0.91 & 0.79 & 0.57 \\
\bottomrule
\end{tabular}
\end{table*}

%% file: figs/auc_sensitivity_table.tex
\begin{table}[t]
\centering
\small
\caption{Sensitivity of model-grounding rankings to AUC parameterisation. Rank correlations are computed against the main configuration.}
\label{tab:auc-sensitivity}
\setlength{\tabcolsep}{2pt}
\begin{tabular}{lccc}
\toprule
\textbf{Configuration} & \textbf{Full-rank $\rho$} & \textbf{Top-10 $\rho$} & \textbf{Top-10 overlap} \\
\midrule
$\tau_{\max}=4$ & 1.00 & 0.95 & 10/10 \\
$\tau_{\max}=6.5$ & 1.00 & 1.00 & 10/10 \\
$\tau_{\max}=8$ & 1.00 & 0.97 & 9/10 \\
$\tau_{\max}=10$ & 1.00 & 0.95 & 9/10 \\
$n_{\mathrm{points}}=15$ & 1.00 & 0.99 & 10/10 \\
$n_{\mathrm{points}}=50$ & 1.00 & 0.99 & 10/10 \\
Linear-$\tau$ integration & 1.00 & 0.96 & 9/10 \\
\bottomrule
\end{tabular}
\end{table}

%% file: figs/model_type_rank_delta_gov-ai_policy.tex
\begin{table*}[t]
\centering
\small
\caption{Median [interquartile] human-grounding rank $-$ MMTEB(Danish) rank per model group.}
\label{tab:model-type-rank-delta}
\begin{tabular}{lrrr}
\toprule
\textbf{Model type} & \textbf{$N$} & \textbf{Policy} & \textbf{Government AI} \\
\midrule
Commercial & 3 & $+2.0$ {\scriptsize [$-10.5$, $+2.0$]} & $-1.0$ {\scriptsize [$-12.5$, $+0.5$]} \\
Instruction-tuned & 4 & $+1.5$ {\scriptsize [$-1.2$, $+7.2$]} & $+1.5$ {\scriptsize [$-3.5$, $+5.5$]} \\
Model2Vec & 10 & $-1.0$ {\scriptsize [$-4.2$, $-1.0$]} & $-3.5$ {\scriptsize [$-6.2$, $+4.8$]} \\
Sentence Transformer & 15 & $+5.0$ {\scriptsize [$-2.5$, $+8.0$]} & $+4.0$ {\scriptsize [$-3.0$, $+7.0$]} \\
\bottomrule
\end{tabular}
\end{table*}

%% file: figs/lexical_baselines_table.tex
\begin{table*}[t]
\centering
\small
\caption{Lexical baseline performance. Ranks are computed within the full set of neural and lexical models.}
\label{tab:lexical-baselines}
\begin{tabular}{llcccc}
\toprule
\textbf{Dataset} & \textbf{Baseline} & \textbf{Alignment AUC} & \textbf{AUC rank} & \textbf{Clustering ARI} & \textbf{ARI rank} \\
\midrule
Responsible AI & \texttt{tfidf-char35} & 0.215 & 23/34 & 0.183 & 3/34 \\
Responsible AI & \texttt{jaccard-binary} & -0.047 & 34/34 & 0.014 & 34/34 \\
Welfare & \texttt{tfidf-char35} & 0.270 & 23/34 & 0.312 & 12/34 \\
Welfare & \texttt{jaccard-binary} & -0.168 & 34/34 & 0.013 & 34/34 \\
Gov-AI & \texttt{tfidf-char35} & 0.151 & 32/34 & 0.103 & 26/34 \\
Gov-AI & \texttt{jaccard-binary} & 0.104 & 34/34 & 0.032 & 34/34 \\
\bottomrule
\end{tabular}
\end{table*}

%% file: figs/respondent_demographics.tex
\begin{table}[ht]
\centering
\caption{Sample composition by author group.}
\label{tab:sample-composition}
\begin{tabular}{lrr}
\toprule
Group & $n$ & \% \\
\midrule
\multicolumn{3}{l}{\textit{Responsible AI}} \\
Female & 170 & 56.5 \\
Male & 123 & 40.9 \\
Unknown & 8 & 2.7 \\
\textbf{Total} & \textbf{301} & \textbf{100.0} \\
\midrule
\multicolumn{3}{l}{\textit{Welfare}} \\
A & 51 & 28.7 \\
B & 46 & 25.8 \\
C & 32 & 18.0 \\
D & 31 & 17.4 \\
E & 18 & 10.1 \\
\textbf{Total} & \textbf{178} & \textbf{100.0} \\
\bottomrule
\end{tabular}
\end{table}

%% file: figs/fairness_triplet_counts.tex
\begin{table}[t]
\centering
\small
\caption{Retained triplet counts by dataset, group, and separation-ratio threshold.}
\label{tab:group-triplet-counts}
\begin{tabular}{llrrr}
\toprule
\textbf{Dataset} & \textbf{Group} & $\mathbf{\tau=1}$ & $\mathbf{\tau=2}$ & $\mathbf{\tau=4}$ \\
\midrule
Responsible AI & Women & 97,988 & 14,523 & 1,956 \\
Responsible AI & Men & 77,984 & 11,560 & 1,522 \\
Welfare & Party 1 & 26,418 & 4,213 & 629 \\
Welfare & Party 11 & 27,041 & 4,159 & 552 \\
Welfare & Party 2 & 26,394 & 3,948 & 495 \\
Welfare & Party 3 & 27,017 & 4,165 & 558 \\
Welfare & Party 4 & 26,919 & 3,920 & 469 \\
\bottomrule
\end{tabular}
\end{table}

%% file: figs/fairness_group_gap_controlled.tex
\begin{table*}[t]
\centering
\small
\setlength{\tabcolsep}{4pt}
\caption{Best--worst demographic group gaps in model-grounding accuracy for the highest-mean-AUC model per dataset, before and after adjustment for lexical and length confounds. See main text for definitions of $\Delta_{\mathrm{group}}$, $\Delta_{\mathrm{adj}}$, and the bootstrap procedure.}
\label{tab:group-gap-controlled}
\begin{tabular}{llcccc}
\toprule
 & & \multicolumn{2}{c}{\textbf{Unadjusted}} & \multicolumn{2}{c}{\textbf{Adjusted}} \\
\cmidrule(lr){3-4} \cmidrule(lr){5-6}
\textbf{Dataset} & \textbf{Model} & $\Delta_{\mathrm{group}}$ [95\% CI] & $p$ & $\Delta_{\mathrm{adj}}$ [95\% CI] & $p$ \\
\midrule
Responsible AI & multilingual-e5-large-instruct & 0.028 [0.026,\,0.029] & <0.001 & 0.011 [0.007,\,0.016] & <0.001 \\
Welfare & paraphrase-multilingual-mpnet-base-v2 & 0.068 [0.062,\,0.073] & <0.001 & 0.031 [0.021,\,0.043] & <0.001 \\
\bottomrule
\end{tabular}
\end{table*}

%% file: figs/statement_coverage_table.tex
\begin{table*}[t]
\centering
\small
\caption{Statement occurrence and co-occurrence statistics by dataset (\secref{sec:data}). Occurrences count the rounds in which a statement appears in any role; co-occurrence is the share of all unordered statement pairs that share at least one round. Rater pairs are the $\binom{n}{2}$ pairs a panel of $n$ participants allows ($N_\text{potential}$) and the subset that co-rated at least one round ($N_\text{sampled}$), which is what between-rater reliability is computed over.}
\label{tab:coverage}
\begin{tabular}{lrrrrrrr}
\toprule
 & & & \multicolumn{2}{c}{Occurrences per statement} & Co-occurring & \multicolumn{2}{c}{Rater pairs} \\
\cmidrule(lr){4-5} \cmidrule(lr){7-8}
Dataset & $N_\text{total}$ & $N_\text{sampled}$ & Mean & Median & \% of pairs & $N_\text{potential}$ & $N_\text{sampled}$ \\
\midrule
Gov-AI & 342 & 244 & 3.44 & 4.00 & 13.07 & 15 & 9 \\
RAI & 873 & 294 & 3.95 & 2.00 & 11.91 & 15 & 8 \\
Welfare & 377 & 174 & 3.45 & 2.00 & 17.77 & 15 & 8 \\
\bottomrule
\end{tabular}
\end{table*}

%% file: latex/appendix_triplets.tex
\section{Qualitative Triplet Analysis}
\label{app:triplets}

We qualitatively analyse model--human triplet comparisons with human separation ratio $r_i(t) > 13$ for the best-performing model (\texttt{paraphrase-multilingual-mpnet}). For each comparison, we show the anchor statement and the statements judged closer to and farther from the anchor by the human rater. The separation ratio $r_i(t)$ describes how unambiguously the human rater expressed this ordering. \textbf{Correct} means that the model preserves the human ordering; \textbf{Incorrect} means that the model reverses it.

\subsection{Responsible AI --- Correct (8 of 24 shown)}
\label{app:triplets-rai-correct}

\paragraph{$r_i(t) = 15.5$}
\begin{description}[style=nextline,leftmargin=1em,font=\normalfont\bfseries]
  \item[Anchor] Kritiske kompetener vil derfor også forsvine.
  \item[Human-judged closer] Dumhed.   Vi mister personlig dømmekraft.   Faglighed er i fremtiden ikke iboende hos individer.   Vi mister kompetencer, da der ikke længere er efterspørgsel på færdigheder og kunnen.
  \item[Human-judged farther] Forskning
\end{description}
\paragraph{$r_i(t) = 15.5$}
\begin{description}[style=nextline,leftmargin=1em,font=\normalfont\bfseries]
  \item[Anchor] Eleverne og studerende lærer kun at bruge 'færdiglavede' løsninger gennem AI, i stedet for at lave sine egne.
  \item[Human-judged closer] Almen dannelse!  Manglende viden om, hvad AI kan bruges til.
  \item[Human-judged farther] Specialiseringa  Ansvarlig AI skal bidrage til at mennesker kan fokusere på at løse problemstillinger, som AI ikke kan hjælpe os med, fx menneskelig relationsarbejde.
\end{description}
\paragraph{$r_i(t) = 15.5$}
\begin{description}[style=nextline,leftmargin=1em,font=\normalfont\bfseries]
  \item[Anchor] AI tilbyder konstruerede sandheder, der kan tilpasses den enkelte behov - hvordan sikrer vi anvendelse af AI, der tilgodeser fællesskabet?
  \item[Human-judged closer] Det er - og bliver for nemt at skabe opdigtet og manipuleret indhold til de forskellige digitale platforme.
  \item[Human-judged farther] Ren hype
\end{description}
\paragraph{$r_i(t) = 15.4$}
\begin{description}[style=nextline,leftmargin=1em,font=\normalfont\bfseries]
  \item[Anchor] Bias og diskrimination  Et program er kun så godt som sin data, og det kan repræsentere et problem hvis det bliver anvendt frit ig uden regulering
  \item[Human-judged closer] AI kan karakterisere os til jobsamtaler, forsikringer og vurdere os som mennesker
  \item[Human-judged farther] Jeg går ud fra at I mener U-ansvarlig.... (slåfejl?)
\end{description}
\paragraph{$r_i(t) = 15.4$}
\begin{description}[style=nextline,leftmargin=1em,font=\normalfont\bfseries]
  \item[Anchor] Uddannelse   På uddannelser er personale ikke klædt på til at tackle AI og kan derfor ikke læse unge mennesker på til at være kritiske ifht materialer og viden skabt via AI.
  \item[Human-judged closer] Uddannelse
  \item[Human-judged farther] Ulighed    Dem som han adgang til den nyeste og kraftigste AI bliver A-holdet.
\end{description}
\paragraph{$r_i(t) = 15.4$}
\begin{description}[style=nextline,leftmargin=1em,font=\normalfont\bfseries]
  \item[Anchor] Algoritme bias  AI-systemer kan indlejre og forstærke bias
  \item[Human-judged closer] AI er ikke klogere end os i fortiden.  Jeg er bekymret for, at AI kun kommer til at forstærke stereotyper, mønstre, og uligheder i vores samfund, når det er baseret på fortiden.
  \item[Human-judged farther] Jeg går ud fra at I mener U-ansvarlig.... (slåfejl?)
\end{description}
\paragraph{$r_i(t) = 15.0$}
\begin{description}[style=nextline,leftmargin=1em,font=\normalfont\bfseries]
  \item[Anchor] Mangel på relevant lovgivning og rel  gulering
  \item[Human-judged closer] Manglende regulering og fokus på at de skal komme alle mennesker til gode uden at ødelægge planeten. Der mangler demokratisk indflydelse fra borgernes side.
  \item[Human-judged farther] Ren hype
\end{description}
\paragraph{$r_i(t) = 14.6$}
\begin{description}[style=nextline,leftmargin=1em,font=\normalfont\bfseries]
  \item[Anchor] Vi er allerede bagud:  Den teknologiske udvikling går hurtigere end hvad det juridiske system kan følge med til.
  \item[Human-judged closer] AI udviklingens fart.  Udviklingen har taget ekstra fart særligt de sidste 10 år selvom, det har været et område, der har eksisteret mange år. Det betyder, at vi er bagud i forhold til mange områder, hvor AI kan anvendes i et omfang, som får behovet for regulering til at stige.
  \item[Human-judged farther] Worst worst case, robotterne tager over og de ser os som skadedyr der skal bekæmpes, lidt ligesom vi sætter myrelokkedåser op (men nok inden for 10 år trods alt)
\end{description}

\subsection{Responsible AI --- Incorrect (8)}
\label{app:triplets-rai-incorrect}

\paragraph{$r_i(t) = 16.0$}
\begin{description}[style=nextline,leftmargin=1em,font=\normalfont\bfseries]
  \item[Anchor] AI tilbyder konstruerede sandheder, der kan tilpasses den enkelte behov - hvordan sikrer vi anvendelse af AI, der tilgodeser fællesskabet?
  \item[Human-judged closer] Det er - og bliver for nemt at skabe opdigtet og manipuleret indhold til de forskellige digitale platforme.
  \item[Human-judged farther] AI er ikke klogere end os i fortiden.  Jeg er bekymret for, at AI kun kommer til at forstærke stereotyper, mønstre, og uligheder i vores samfund, når det er baseret på fortiden.
\end{description}
\paragraph{$r_i(t) = 14.5$}
\begin{description}[style=nextline,leftmargin=1em,font=\normalfont\bfseries]
  \item[Anchor] Almen dannelse!  Manglende viden om, hvad AI kan bruges til.
  \item[Human-judged closer] Eleverne og studerende lærer kun at bruge 'færdiglavede' løsninger gennem AI, i stedet for at lave sine egne.
  \item[Human-judged farther] Specialiseringa  Ansvarlig AI skal bidrage til at mennesker kan fokusere på at løse problemstillinger, som AI ikke kan hjælpe os med, fx menneskelig relationsarbejde.
\end{description}
\paragraph{$r_i(t) = 13.8$}
\begin{description}[style=nextline,leftmargin=1em,font=\normalfont\bfseries]
  \item[Anchor] Psykiatrien
  \item[Human-judged closer] Vi har brug for et samfund med omsorg og nærvær - det kan vi ikke få digitalt
  \item[Human-judged farther] Forskning
\end{description}
\paragraph{$r_i(t) = 13.6$}
\begin{description}[style=nextline,leftmargin=1em,font=\normalfont\bfseries]
  \item[Anchor] Kritiske kompetener vil derfor også forsvine.
  \item[Human-judged closer] Vi har et skolesystem og velfærdsystem i knæ, det kan vi ikke løse digitalt
  \item[Human-judged farther] Økonomiske interesser:  Stærke økonomiske interesser kan let stå i vejen for passende politisk regulering.
\end{description}
\paragraph{$r_i(t) = 13.4$}
\begin{description}[style=nextline,leftmargin=1em,font=\normalfont\bfseries]
  \item[Anchor] Dumhed.   Vi mister personlig dømmekraft.   Faglighed er i fremtiden ikke iboende hos individer.   Vi mister kompetencer, da der ikke længere er efterspørgsel på færdigheder og kunnen.
  \item[Human-judged closer] Ukontrolleret anvendelse af AI
  \item[Human-judged farther] Mener ikke det er et problem hvis der er tale om ansvarlig udvikling og anvendelse.
\end{description}
\paragraph{$r_i(t) = 13.3$}
\begin{description}[style=nextline,leftmargin=1em,font=\normalfont\bfseries]
  \item[Anchor] Facttjekke  Med al misinformationen i medierne i dag kan det blive svært at facttjekke, hvis meget indhold generet af AI og især hvis dette er bareseret på hinanden
  \item[Human-judged closer] Det er - og bliver for nemt at skabe opdigtet og manipuleret indhold til de forskellige digitale platforme.
  \item[Human-judged farther] AI er ikke klogere end os i fortiden.  Jeg er bekymret for, at AI kun kommer til at forstærke stereotyper, mønstre, og uligheder i vores samfund, når det er baseret på fortiden.
\end{description}
\paragraph{$r_i(t) = 13.3$}
\begin{description}[style=nextline,leftmargin=1em,font=\normalfont\bfseries]
  \item[Anchor] Manglende gennemsigtighed
  \item[Human-judged closer] Tillid:  Hvordan og af hvem trænes AI til at oparbejde, forstå og indeholde menneskelige egenskaber?
  \item[Human-judged farther] Manglende etiske retningslinjer og regulering  Udviklingen sker så hurtig, at lovgivning vil have svært ved at følge med
\end{description}
\paragraph{$r_i(t) = 13.1$}
\begin{description}[style=nextline,leftmargin=1em,font=\normalfont\bfseries]
  \item[Anchor] Penge  Der er rigtig mange penge involveret i at bruge data til at reklamere osv overfor danskere.
  \item[Human-judged closer] IT-giganter, vi ikke kan tøjle    Vi kan ikke styre IT-giganter som google, meta, microsoft og chatGPT. Det bliver meget svært at regulere det her område selv.
  \item[Human-judged farther] Ren hype
\end{description}

\subsection{Welfare --- Correct (3 of 16 shown)}
\label{app:triplets-welfare-correct}

\paragraph{$r_i(t) = 16.2$}
\begin{description}[style=nextline,leftmargin=1em,font=\normalfont\bfseries]
  \item[Anchor] Demografi.  Massive ændringer i befolkningssammensætningen vil betyde øgede udgifter med mange ældre med behov for pleje og sundhedsydelser.
  \item[Human-judged closer] Demografi  Meget flere ældre end unge
  \item[Human-judged farther] Øgede forsvarsudgifter
\end{description}
\paragraph{$r_i(t) = 14.4$}
\begin{description}[style=nextline,leftmargin=1em,font=\normalfont\bfseries]
  \item[Anchor] Den øgede privatisering har skabt et a- og b-hold, hvor a-holdet har deres på det tørre gennem sundhedsforsikringer osv - og derfor er det dem ikke så magtpåliggende at styrke velfærden for samfundets svageste.
  \item[Human-judged closer] De privatpraktiserende læger har for stor magt, og deres flaskehalsfunktion er dobbeltrettet, for selv om den holder folk ude af sygehusene, så låser den også kommunen i gammeldags serviceniveauer.
  \item[Human-judged farther] Den nye sikkerheds situation og investeringer i militæret.
\end{description}
\paragraph{$r_i(t) = 13.0$}
\begin{description}[style=nextline,leftmargin=1em,font=\normalfont\bfseries]
  \item[Anchor] Kompliceret og omfattende central regulering   Eksisterende lovgivning og flere reformer gøre kommunalpolitikerne de facto til en slags embedsmænd, der blot skal forvalte folketingets retningslinjer. Kommunerne mangler frier rammer for at lave lokale løsninger.
  \item[Human-judged closer] Det tager for lang tid at revidere budgetloven
  \item[Human-judged farther] Social ulighed og polarisering
\end{description}

\subsection{Welfare --- Incorrect (3)}
\label{app:triplets-welfare-incorrect}

\paragraph{$r_i(t) = 13.5$}
\begin{description}[style=nextline,leftmargin=1em,font=\normalfont\bfseries]
  \item[Anchor] Demografi.  Massive ændringer i befolkningssammensætningen vil betyde øgede udgifter med mange ældre med behov for pleje og sundhedsydelser.
  \item[Human-judged closer] Demografi  Meget flere ældre end unge
  \item[Human-judged farther] Der kommer flere og flere komplekse opgaver ud til den kommunale pleje især med ønsket om at løfte flere opgaver i det nære sundhedsområde.  Hvis ikke ressourcer er til stede, kan opgaverne blive svære at løfte
\end{description}
\paragraph{$r_i(t) = 13.4$}
\begin{description}[style=nextline,leftmargin=1em,font=\normalfont\bfseries]
  \item[Anchor] Sikring mod klimaforandringer
  \item[Human-judged closer] At grøn trepart går amok , som minkerstatningerne
  \item[Human-judged farther] Destruktive angreb udefra på vores fysiske og teknologiske infrastrukturer
\end{description}
\paragraph{$r_i(t) = 13.0$}
\begin{description}[style=nextline,leftmargin=1em,font=\normalfont\bfseries]
  \item[Anchor] For stort og tungt et administrativt system
  \item[Human-judged closer] Prioritering af andre indsatsområder (Forsvaret, skattelettelser, klima mv.) fra Christiansborg. Derfor nedprioritering af kommunernes økonomi til den borgernære velfærd.
  \item[Human-judged farther] For lille produktivitet
\end{description}

\subsection{Government AI --- Correct (10 of 82 shown)}
\label{app:triplets-gov-ai-correct}

\paragraph{$r_i(t) = 69.9$}
\begin{description}[style=nextline,leftmargin=1em,font=\normalfont\bfseries]
  \item[Anchor] Use Case: Avian population estimates from passive acoustic monitoring
Description: Reliable estimates of avian abundance from acoustic recordings
  \item[Human-judged closer] Use Case: Liable Party Research
Description: Supervised machine learning model for scraping leasing documents for current liable party ownership.
  \item[Human-judged farther] Use Case: Email Prioritization in Outlook
Description: The Outlook in Office365 suite can categorize incoming emails based on users' definitions, which can help users manage their inbox more efficiently. The output is categorized emails.
\end{description}
\paragraph{$r_i(t) = 69.5$}
\begin{description}[style=nextline,leftmargin=1em,font=\normalfont\bfseries]
  \item[Anchor] Use Case: Lava lake thermal pattern classification using self organizing maps and relationships to eruption pr
Description: We apply a machine learning algorithm called self-organizing maps (SOM) to thermal infrared time-lapse images .
  \item[Human-judged closer] Use Case: Analytical Chemistry \& Geologic Resources Identification from X-rays
Description:   Determining chemistry and rock composition from X-ray-based analytical chemistry platforms
  \item[Human-judged farther] Use Case: Agency Support Companion (ASC) chatbot
Description: Generative AI model to create content, summarization and research capabilities
\end{description}
\paragraph{$r_i(t) = 45.2$}
\begin{description}[style=nextline,leftmargin=1em,font=\normalfont\bfseries]
  \item[Anchor] Use Case: AWS AI Testbed
Description: Purpose: Components test AWS AI tools in a test environment to assess utility. 

Expected benefits: Improves efficiencies, productivity, and data quality.
  \item[Human-judged closer] Use Case: Source Code Development Tool
Description: This capability will enable end users to develop software faster and more efficiently through the use of a generative artificial intelligence coding assistant.  

  \item[Human-judged farther] Use Case: VA Chat Copilot Meta Pilot
Description: Administrative efficiency for VA employees 
\end{description}
\paragraph{$r_i(t) = 40.4$}
\begin{description}[style=nextline,leftmargin=1em,font=\normalfont\bfseries]
  \item[Anchor] Use Case: Agency Support Companion (ASC) chatbot
Description: Generative AI model to create content, summarization and research capabilities
  \item[Human-judged closer] Use Case: CFOO Travel Chatbot
Description: Compiling data from internal FDIC website and targeted documents to answer questions about FDIC travel. Makes information more readily available, searchable and readable.
  \item[Human-judged farther] Use Case: Everglades-Flux, Digital Surveys
Description: We are creating a program that can automatically process Normalized Difference Vegetation Index images and come up with a true value of live vegetation. It will also gap fill missing data via AI/ML programs.
\end{description}
\paragraph{$r_i(t) = 36.4$}
\begin{description}[style=nextline,leftmargin=1em,font=\normalfont\bfseries]
  \item[Anchor] Use Case: Data modeling and visualization by MS Excel
Description: MS Excel contains many data analysis toolkits to provide AI-powered data analysis and visualization, which can improve understanding and presentation of data. 
  \item[Human-judged closer] Use Case: FOIA processing
Description: No tool has been selected yet, exploring tools tht may help with sorting requests
  \item[Human-judged farther] Use Case: Semantic Similarity Text Search
Description: Semantic search of agency library of safety recommendation and report texts, improving speed and efficiency of researching previously identified transportation safety topics
\end{description}
\paragraph{$r_i(t) = 36.4$}
\begin{description}[style=nextline,leftmargin=1em,font=\normalfont\bfseries]
  \item[Anchor] Use Case: CFOO Travel Chatbot
Description: Compiling data from internal FDIC website and targeted documents to answer questions about FDIC travel. Makes information more readily available, searchable and readable.
  \item[Human-judged closer] Use Case: Patent Automatic Document Code Determination
Description: Quality control of customer applied document labels
  \item[Human-judged farther] Use Case: BlueTeam AI
Description: BlueTeam AI enables network monitoring capabilities to provide governance and guardrails for LLM Generative AI use for the enterprise.
\end{description}
\paragraph{$r_i(t) = 32.3$}
\begin{description}[style=nextline,leftmargin=1em,font=\normalfont\bfseries]
  \item[Anchor] Use Case: Using machine learning to detect invasive bullfrogs
Description: Detecting bullfrogs along their invasion front in order to inform removal efforts
  \item[Human-judged closer] Use Case: Shoreline Modeling
Description: Exploring the prospect of using AI/ML models to predict shoreline change and compare accuracy to traditional models. Examples of AI/ML models include LSTM, CNN, and Transformers
  \item[Human-judged farther] Use Case: News from commercial publisher such as Google News
Description: Major news publishers, such as Google News, send selective news based on users' preferences and recent visits. 
\end{description}
\paragraph{$r_i(t) = 27.8$}
\begin{description}[style=nextline,leftmargin=1em,font=\normalfont\bfseries]
  \item[Anchor] Use Case: DOL Intranet Website Chatbot Assistant
Description: 
Conversational AI Assistant \& DOL intranet websites to help answer common procurement questions, as well as specific contract questions.
  \item[Human-judged closer] Use Case: Agency Support Companion (ASC) chatbot
Description: Generative AI model to create content, summarization and research capabilities
  \item[Human-judged farther] Use Case: Qualitative analysis
Description: ACF staff often conduct surveys and interviews, which generate qualitative data that needs to be analyzed for themes and trends.

The AI functionality assists users in identifying relevant passages in their narrative data related to specific topics.
\end{description}
\paragraph{$r_i(t) = 26.5$}
\begin{description}[style=nextline,leftmargin=1em,font=\normalfont\bfseries]
  \item[Anchor] Use Case: Semantic Similarity Text Search
Description: Semantic search of agency library of safety recommendation and report texts, improving speed and efficiency of researching previously identified transportation safety topics
  \item[Human-judged closer] Use Case: National Training Team | Schools  FAQ Dashboard
Description: The model should classify thousands of questions into categories, generate a sample question based on the category, and provide a suggested answer based on previous responses to similar questions.
  \item[Human-judged farther] Use Case: Qualitative analysis
Description: ACF staff often conduct surveys and interviews, which generate qualitative data that needs to be analyzed for themes and trends.

The AI functionality assists users in identifying relevant passages in their narrative data related to specific topics.
\end{description}
\paragraph{$r_i(t) = 26.1$}
\begin{description}[style=nextline,leftmargin=1em,font=\normalfont\bfseries]
  \item[Anchor] Use Case: Analytical Chemistry \& Geologic Resources Identification from X-rays
Description:   Determining chemistry and rock composition from X-ray-based analytical chemistry platforms
  \item[Human-judged closer] Use Case: Everglades-Flux, Digital Surveys
Description: We are creating a program that can automatically process Normalized Difference Vegetation Index images and come up with a true value of live vegetation. It will also gap fill missing data via AI/ML programs.
  \item[Human-judged farther] Use Case: Agency Support Companion (ASC) chatbot
Description: Generative AI model to create content, summarization and research capabilities
\end{description}

\subsection{Government AI --- Incorrect (10 of 54 shown)}
\label{app:triplets-gov-ai-incorrect}

\paragraph{$r_i(t) = 70.1$}
\begin{description}[style=nextline,leftmargin=1em,font=\normalfont\bfseries]
  \item[Anchor] Use Case: Liable Party Research
Description: Supervised machine learning model for scraping leasing documents for current liable party ownership.
  \item[Human-judged closer] Use Case: Avian population estimates from passive acoustic monitoring
Description: Reliable estimates of avian abundance from acoustic recordings
  \item[Human-judged farther] Use Case: Email Prioritization in Outlook
Description: The Outlook in Office365 suite can categorize incoming emails based on users' definitions, which can help users manage their inbox more efficiently. The output is categorized emails.
\end{description}
\paragraph{$r_i(t) = 45.4$}
\begin{description}[style=nextline,leftmargin=1em,font=\normalfont\bfseries]
  \item[Anchor] Use Case: AWS AI Testbed
Description: Purpose: Components test AWS AI tools in a test environment to assess utility. 

Expected benefits: Improves efficiencies, productivity, and data quality.
  \item[Human-judged closer] Use Case: Source Code Development Tool
Description: This capability will enable end users to develop software faster and more efficiently through the use of a generative artificial intelligence coding assistant.  

  \item[Human-judged farther] Use Case: DOIChatGPT API Management Development Environment
Description:  This will be a public facing version of the DOIChatGPT APIM and Azure OpenAI that is used for training and proof of concept validation (with sanitized data) prior to migrating to production DOIChatGPT APIM.
\end{description}
\paragraph{$r_i(t) = 36.6$}
\begin{description}[style=nextline,leftmargin=1em,font=\normalfont\bfseries]
  \item[Anchor] Use Case: Data modeling and visualization by MS Excel
Description: MS Excel contains many data analysis toolkits to provide AI-powered data analysis and visualization, which can improve understanding and presentation of data. 
  \item[Human-judged closer] Use Case: FOIA processing
Description: No tool has been selected yet, exploring tools tht may help with sorting requests
  \item[Human-judged farther] Use Case: National Training Team | Schools  FAQ Dashboard
Description: The model should classify thousands of questions into categories, generate a sample question based on the category, and provide a suggested answer based on previous responses to similar questions.
\end{description}
\paragraph{$r_i(t) = 35.4$}
\begin{description}[style=nextline,leftmargin=1em,font=\normalfont\bfseries]
  \item[Anchor] Use Case: CFOO Travel Chatbot
Description: Compiling data from internal FDIC website and targeted documents to answer questions about FDIC travel. Makes information more readily available, searchable and readable.
  \item[Human-judged closer] Use Case: Patent Automatic Document Code Determination
Description: Quality control of customer applied document labels
  \item[Human-judged farther] Use Case: Tableau Public 2023
Description: Allows users to share data visualizations, providing interactive insights.
\end{description}
\paragraph{$r_i(t) = 32.0$}
\begin{description}[style=nextline,leftmargin=1em,font=\normalfont\bfseries]
  \item[Anchor] Use Case: Shoreline Modeling
Description: Exploring the prospect of using AI/ML models to predict shoreline change and compare accuracy to traditional models. Examples of AI/ML models include LSTM, CNN, and Transformers
  \item[Human-judged closer] Use Case: Using machine learning to detect invasive bullfrogs
Description: Detecting bullfrogs along their invasion front in order to inform removal efforts
  \item[Human-judged farther] Use Case: News from commercial publisher such as Google News
Description: Major news publishers, such as Google News, send selective news based on users' preferences and recent visits. 
\end{description}
\paragraph{$r_i(t) = 30.3$}
\begin{description}[style=nextline,leftmargin=1em,font=\normalfont\bfseries]
  \item[Anchor] Use Case: CFOO Travel Chatbot
Description: Compiling data from internal FDIC website and targeted documents to answer questions about FDIC travel. Makes information more readily available, searchable and readable.
  \item[Human-judged closer] Use Case: Agency Support Companion (ASC) chatbot
Description: Generative AI model to create content, summarization and research capabilities
  \item[Human-judged farther] Use Case: FDIC Financial Audit Report
Description: Monitor employee transactions ensure regulations and policies are being followed. Reduce time to receive answers, allowing folks to focus on more critical mission-driven work.
\end{description}
\paragraph{$r_i(t) = 30.1$}
\begin{description}[style=nextline,leftmargin=1em,font=\normalfont\bfseries]
  \item[Anchor] Use Case: Deep-learning Integration into NEIC Operations
Description: The National Earthquake Information Center (NEIC) is improving its earthquake detection and processing systems by leveraging artificial intelligence and machine learning.
  \item[Human-judged closer] Use Case: Position Description Generation and Evaluation
Description: Faster and more accurate creation of PDs, reducing admin burden on field and HQs as well as HR specialists, allowing agency to get more done with less staff.
  \item[Human-judged farther] Use Case: AI Chat Bot for IT User Services
Description: Interactive RAG chat bot model trained on existing, updated and new IT resource documentation in the user space. Platform will be used as an informative method for users to handle tier 1 IT issues and help guide to the correct place.
\end{description}
\paragraph{$r_i(t) = 28.8$}
\begin{description}[style=nextline,leftmargin=1em,font=\normalfont\bfseries]
  \item[Anchor] Use Case: Agency Support Companion (ASC) chatbot
Description: Generative AI model to create content, summarization and research capabilities
  \item[Human-judged closer] Use Case: DOL Intranet Website Chatbot Assistant
Description: 
Conversational AI Assistant \& DOL intranet websites to help answer common procurement questions, as well as specific contract questions.
  \item[Human-judged farther] Use Case: Qualitative analysis
Description: ACF staff often conduct surveys and interviews, which generate qualitative data that needs to be analyzed for themes and trends.

The AI functionality assists users in identifying relevant passages in their narrative data related to specific topics.
\end{description}
\paragraph{$r_i(t) = 26.7$}
\begin{description}[style=nextline,leftmargin=1em,font=\normalfont\bfseries]
  \item[Anchor] Use Case: National Training Team | Schools  FAQ Dashboard
Description: The model should classify thousands of questions into categories, generate a sample question based on the category, and provide a suggested answer based on previous responses to similar questions.
  \item[Human-judged closer] Use Case: Semantic Similarity Text Search
Description: Semantic search of agency library of safety recommendation and report texts, improving speed and efficiency of researching previously identified transportation safety topics
  \item[Human-judged farther] Use Case: Qualitative analysis
Description: ACF staff often conduct surveys and interviews, which generate qualitative data that needs to be analyzed for themes and trends.

The AI functionality assists users in identifying relevant passages in their narrative data related to specific topics.
\end{description}
\paragraph{$r_i(t) = 26.6$}
\begin{description}[style=nextline,leftmargin=1em,font=\normalfont\bfseries]
  \item[Anchor] Use Case: Everglades-Flux, Digital Surveys
Description: We are creating a program that can automatically process Normalized Difference Vegetation Index images and come up with a true value of live vegetation. It will also gap fill missing data via AI/ML programs.
  \item[Human-judged closer] Use Case: Analytical Chemistry \& Geologic Resources Identification from X-rays
Description:   Determining chemistry and rock composition from X-ray-based analytical chemistry platforms
  \item[Human-judged farther] Use Case: Agency Support Companion (ASC) chatbot
Description: Generative AI model to create content, summarization and research capabilities
\end{description}